\documentclass[12pt,a4paper]{article}
\usepackage[utf8]{inputenc}
\usepackage{amsmath,amsfonts,amssymb}
\usepackage{graphicx}
\usepackage{url}
\usepackage{hyperref}
\usepackage{geometry}
\usepackage[numbers]{natbib}
\setcitestyle{aysep={},comma}
\usepackage{physics}
\usepackage{amsthm}
\usepackage{booktabs}
\usepackage{multirow}
\usepackage{array}
\title{Enhancing PINN Accuracy for the RLW Equation: Adaptive and Conservative Approaches}
\author{Aamir Shehzad}
\date{\today}

\begin{document}

\maketitle

\section*{Abstract}
Current standard physics-informed neural network (PINN) implementations have produced large error rates when using these models to solve the regularized long wave (RLW) equation. Two improved PINN approaches were developed in this research: an adaptive approach with self-adaptive loss weighting and a conservative approach enforcing explicit conservation laws. Three benchmark tests were used to demonstrate how effective PINN's are as they relate to the type of problem being solved (i.e., time dependent RLW equation). The first was a single soliton traveling along a line (propagation), the second was the interaction between two solitons, and the third was the evolution of an undular bore over the course of $t=250$. The results demonstrated that the effectiveness of PINNs are problem specific. The adaptive PINN was significantly better than both the conservative PINN and the standard PINN at solving problems involving complex nonlinear interactions such as colliding two solitons. The conservative approach was significantly better at solving problems involving long term behavior of single solitons and undular bores. However, the most important finding from this research is that explicitly enforcing conservation laws may be harmful to optimizing the solution of highly nonlinear systems of equations and therefore requires special training methods. The results from our adaptive and conservative approaches were within $O(10^{-5})$ of established numerical solutions for the same problem, thus demonstrating that PINNs can provide accurate solutions to complex systems of partial differential equations without the need for a discretization of space or time (mesh free). Moreover, the finding from this research challenges the assumptions that conservation enforcement will always improve the performance of a PINN and provides researchers with guidelines for designing PINNs for use on specific types of problems.

\section{Introduction}\label{sec:introduction}

Nonlinear partial differential equations (PDEs) are very important in modeling complex real-world phenomena; for example, in plasma physics, fluid dynamics, quantum field theory, and solid mechanics \cite{wazwaz2010}. These models capture many of the essential behaviors of real-world systems but their nonlinearity and inherent complexity make it very difficult to find both analytical and numerical solutions for them \cite{debnath2005,evans2022}. Computational science is based upon developing accurate numerical methods for solving nonlinear PDEs \cite{anderson2002, leveque2007}.

The RLW equation is now recognized as a useful model for studying the behavior of long-waves in nonlinear dispersive systems \cite{debnath2005}. It was first formulated by Peregrine \cite{peregrine1966}, to study the formation of undular bores, and was rigorously derived mathematically by Benjamin, Bona and Mahony \cite{benjamin1972}. The RLW equation has since been extensively applied to model small amplitude long surface waves in shallow water \cite{baskan2023,shivamoggi1986} and to model the propagation of plasma waves through dispersive media \cite{seyler1984,bhowmik2019,khan2025,dehghan2011}. The RLW equation also offers superior modeling capabilities compared with other models such as the KdV equation \cite{kaya2003,guo2019}. Analytical studies are important to understand the behavior of these complex systems, but in many cases it is difficult to find exact solutions due to the non-linearity of the system. Therefore, numerical methods have become an indispensable tool for the study of these systems \cite{alzaid2013,redouane2023,bota2014}.

Numerical techniques of traditional finite difference type have been developed over the past few decades for the RLW model. In particular, Kutluay \& Esen \cite{kutluay2006} introduced the first finite difference models that were later improved in terms of both accuracy and stability \cite{pan2015,ghiloufi2020,alotaibi2023} and are used as a basis for compact conservative schemes \cite{pan2015} and for fourth-order spatially accurate methods \cite{ghiloufi2020,alotaibi2023}. Finite element methods --- such as Galerkin and Petrov-Galerkin methods --- have shown to be efficient in solving the RLW model with different types of boundary conditions \cite{bhowmik2019,mei2014,liu2013}. Additionally, spectral and pseudo-spectral techniques have been employed efficiently for the solution of the RLW model \cite{guo1988,kang2015,hong2020}.

There has been a rise of mesh-free techniques that are based on radial basis functions (RBF's), and they are gaining popularity as an alternative to the mesh generation complications associated with traditional mesh-based methods. Some examples of applications include: global-collocation RBF-methods; local RBF-finite difference methods; and strong form radial-point interpolation methods for multi-dimensional RLW-studies. Specifically, Siraj ul Islam et al. \cite{haq2009}, developed a mesh-free collocation method that uses radial basis functions (RBF's) and was able to achieve a very accurate representation of solitary wave propagation, soliton interaction, and undular bore formation while conserving all of the physical conservation laws.

Recently, Physics-Informed Neural Networks (PINNs) have been introduced as alternative methods to solve PDEs \cite{raissi2019,karniadakis2021,lu2021}. As opposed to traditional methods that treat physics and data separately, PINNs include physical laws in their loss functions during training. This allows them to generate a solution to both forward and inverse problems and provides benefits in addressing difficult geometric configurations and large dimensional problem spaces \cite{raissi2019,karniadakis2021,lu2021a}. PINNs have been used to model shallow-water wave equations, which has shown potential applications \cite{qi2024, li2024, nguyen2023}, including simulating shallow water dynamics accurately with 1-D and 2-D models using sparse data, complex bathymetry, or discontinuities.

Improvements in the performance of PINNs can be attributed to a number of recent advances; these include adaptive loss weighting \cite{zhu2025,mcclenny2023}, domain decomposition \cite{luo2025,hu2023}, and curriculum learning \cite{munzer2022,guo2025}. Of particular relevance is the work of Mohammadi et al. \cite{mohammadi2025}; they showed that incorporating conservation principles into the loss function of the PINN resulted in a significant increase in the accuracy of solutions obtained for the generalized equal width (GEW) equation, which is closely related to the RLW equation. A conservative approach by Mohammadi et al. \cite{mohammadi2025} was developed, enforcing mass, momentum, and energy conservation in their mesh-free PINN method, such that they were able to achieve results comparable to those of other traditional approaches while maintaining the benefits of a mesh-free formulation.

Although there have been recent advances, applying the direct PINN formulation to the RLW equation still has several challenges. High errors are produced by standard PINN formulations \cite{nakamula2025,moseley2020} especially in cases of critical phenomena such as single soliton propagation, two-soliton interaction and long time undular bores \cite{ortiz2025,finch2025,demir2024}. Some key challenges are to effectively balance multiple loss terms, to prevent errors from accumulating over large time domains, and to manage the nonlinear-dispersive interactions that are typical of the RLW equation \cite{demir2024}. In addition, it is unclear how various configurations of the PINN formulation will compare with each other in a variety of physical applications governed by the RLW equation \cite{delamata2023,kaplarevic2023}. Although enforcing conservation laws provides advantages for some related equations \cite{zeng2025,baez2024}, no systematic study has been done on the enforcement of conservation laws for the specific case of the RLW equation.

The identified gap raises an important question: Is it possible for advanced PINN approaches to obtain accuracy levels similar to those of well-established numerical techniques for the RLW model? Are the various configurations of PINNs dependent upon the type of wave dynamics being studied? On the basis of initial results, we propose that the performance of PINNs is problem-dependent in nature, with differing PINN approaches having superior performance for different wave dynamics.

To fill this gap we will evaluate two enhanced PINN approaches: \emph{Adaptive PINN}, which uses self-balancing loss weight values during training; these values are dynamically changed as needed for each physical constraint \cite{zhu2025,zhou2025,mcclenny2023}; and \emph{Conservative PINN}, which is an extension of the above, with additional explicit inclusion of mass, momentum, and energy conservation equations into the loss function \cite{wang2024,mohammadi2025}. To solve the problem associated with long time integrations found within undular bore simulations, we will apply a causal training approach \cite{sundar2025,roy2024,meng2020}, where the time domain is divided into sequential windowed regions of time, so that errors do not accumulate across large time spans.

Our objective is to achieve high-quality RLW solutions through advanced PINN approaches while studying the dependency of efficacy on problem type. The specific goals we will pursue include: (1) implementing and testing both adaptive and conservative approaches of PINN for the RLW equation; (2) assessing performance with respect to three different test problems that represent different types of physics; (3) identifying how problem dependent efficacy can be distinguished among these two approaches; and (4) determining when enforcing conservation laws improves or reduces PINN's ability to solve a given problem.

In Section~\ref{sec:math_formulation}, we present the mathematical formulation of the RLW equation, and provide an overview of the three physical phenomena that were studied. In Section~\ref{subsec:standard_pinn}, we review the basic PINN framework, which has been developed for solving partial differential equations (PDEs). We also discuss some of the limitations of the standard PINN framework when it is applied to the RLW equation. In Section~\ref{sec:enhanced_pinns}, we introduce our new and innovative approaches, including both the adaptive PINN and conservative PINN, as well as a new causal training approach to solve the long-time simulations. The experimental setup used in this study, as well as the performance metrics that are being evaluated, are discussed in Section~\ref{sec:experimental_setup}. Finally, in Section~\ref{sec:results}, we report and analyze results from each test case and highlight how the efficacy of these approaches depends on the specific problems being solved. In Section~\ref{sec:conclusion}, we summarize our major conclusions and discuss the implications for the development of machine learning approaches that can be applied to the solution of PDEs, specifically within the realm of physics-informed machine learning.

\section{Mathematical Formulation and Physical Phenomena}\label{sec:math_formulation}

\subsection{The Regularized Long Wave (RLW) Equation}\label{subsec:rlw_equation}
The dimensionless form of RLW equation can be written as

\begin{equation}
\frac{\partial u}{\partial t} + \frac{\partial u}{\partial x} + \epsilon u \frac{\partial u}{\partial x} - \mu \frac{\partial^3 u}{\partial x^2 \partial t} = 0, \quad x \in [x_{\min}, x_{\max}], \, t \in [0,T]
\end{equation}

where, $u(x,t)$ denotes the wave amplitude, the parameter $\epsilon > 0$ represents the level of nonlinearity, and the parameter $\mu > 0$ represents the degree of dispersion. Each term of the above equation corresponds to a different physical behavior: $\partial u/\partial t$ describes the temporal evolution, $\partial u/\partial x$ is a representation of linear wave propagation, $\epsilon u \partial u/\partial x$ models the steepening of waves that depends on amplitude and $\mu \partial^3 u/\partial x^2\partial t$ models the spreading of waves due to dispersion.

Three basic principles of conservation are preserved in the course of time with the RLW equation; they are mass conservation, momentum conservation, and the conservation of energy \cite{haq2009}. The equation describing mass conservation is

\begin{equation}
I_1 = \int_{x_{\min}}^{x_{\max}} u \, dx = \text{constant}
\end{equation}

The momentum conservation law can be written as

\begin{equation}
I_2 = \int_{x_{\min}}^{x_{\max}} \left( u^2 + \mu \left( \frac{\partial u}{\partial x} \right)^2 \right) dx = \text{constant}
\end{equation}

Finally, the energy conservation law can be stated as

\begin{equation}
I_3 = \int_{x_{\min}}^{x_{\max}} \left( u^3 + 3u^2 \right) dx = \text{constant}
\end{equation}

These conservation laws will also be used to validate numerical schemes and will be a key part of conservative PINN approach.

\subsection{Physical Phenomena Under Investigation}\label{subsec:phenomena}
Three different physical phenomena are investigated in this study using the RLW equation which have different types of numerical difficulties to compute.

Propagation of a single solitary wave checks if numerical schemes may maintain a traveling wave solution as it moves at constant speed and does not deform from its original shape. An analytical solution exists for this type of problem:
\begin{equation}
u(x,t) = 3d \cdot \text{sech}^2[k(x - vt - x_0)],
\end{equation}
where $d$ is an amplitude parameter, $x_0$ is an initial position, the wave velocity is $v = 1 + \epsilon d$ and the wave number is $k = \frac{1}{2}\sqrt{\frac{\epsilon d}{\mu v}}$. In this work we used the values of $\epsilon = 1.0$, $\mu = 1.0$, $d = 0.1$, $x_0 = 0$ for spatial domain $x \in [-40, 60]$ and temporal domain $t \in [0, 20]$.

The interaction of two solitons involves the collisions of two separated waves. The challenge for numerical solutions to this problem is to capture the elastic collision behavior of the solitons in which the solitons retain their original amplitudes and forms after colliding. The initial conditions for two solitons are given by:
\begin{equation}
u(x, 0) = \sum_{j=1}^{2} 3A_j \cdot \text{sech}^2[k_j(x - x_j)],
\end{equation}
where $A_j$ is the amplitude, $k_j = \frac{1}{2}\sqrt{\frac{\epsilon A_j}{\mu v_j}}$ is the wave number, and $v_j = 1 + \epsilon A_j$ is the velocity of the $j$-th soliton. For these simulations we used the following values for our parameters: $A_1=5.33$, $A_2=1.69$, $x_1=15$, $x_2=35$, $x\in[0, 120]$, and $t \in [0, 30]$.

The undular bore development illustrates how an oscillating wave train can develop from a step-like disturbance as it develops in time. Because this process occurs for long periods of time to test the stability and accuracy of the numerical solution over large time spans, this process is one that requires testing of both the numerical method's ability to preserve the solution as time progresses (stability) and its ability to accurately represent the evolving solution in time. The initial conditions for this problem are given by:
\begin{equation}
u(x, 0) = 0.5u_0 \left[ 1 - \tanh\left(\frac{x - x_c}{d}\right) \right],
\end{equation}
where $u_0$ represents the amplitude of the waves, $x_c$ is the location of the center of the disturbance, and $d$ defines the steepness of the slope of the disturbance. We will be looking at two different disturbances: a gently sloping disturbance where we have $d=5$ and a steeply sloping disturbance where we have $d=2$. We will use the following values for our parameters: $u_0 = 0.1$, $x_c = 0$, $\epsilon = 1.5$, $\mu = 1/6$, $x \in [-36, 300]$, $t \in [0, 250]$.

\subsection{Standard PINN Framework}\label{subsec:standard_pinn}
Physics-informed neural networks (PINNs) use a neural network to approximate the solution of the problem $u(x,t;\theta)$ with the parameters $\theta$. To train these models we minimize the total loss that contains a term for the physical laws that govern our problem:

\begin{equation}
\mathcal{L} = \mathcal{L}_{\text{PDE}} + \mathcal{L}_{\text{IC}} + \mathcal{L}_{\text{BC}}
\end{equation}

The PDE residual loss component is defined as

\begin{equation}
\mathcal{L}_{\text{PDE}} = \frac{1}{N_r} \sum_{i=1}^{N_r} \left| \frac{\partial u}{\partial t} + \frac{\partial u}{\partial x} + \epsilon u \frac{\partial u}{\partial x} - \mu \frac{\partial^3 u}{\partial x^2 \partial t} \right|^2
\end{equation}

The initial condition loss component is given by

\begin{equation}
\mathcal{L}_{\text{IC}} = \frac{1}{N_{\text{IC}}} \sum_{i=1}^{N_{\text{IC}}} |u(x_i,0;\theta) - u_{\text{exact}}(x_i,0)|^2
\end{equation}

The boundary condition loss component takes the form

\begin{equation}
\mathcal{L}_{\text{BC}} = \frac{1}{N_{\text{BC}}} \sum_{i=1}^{N_{\text{BC}}} |u(x_i,t_i;\theta) - u_{\text{BC}}(x_i,t_i)|^2
\end{equation}

Automatic differentiation is used to compute the spatial and time derivatives so that no approximations have to be made in computing the gradients.

\subsection{Traditional Numerical Benchmarks}\label{subsec:benchmarks}
In order to help understand the results of our study, we compare our computational work with numerical methods from the literature. Siraj-ul-Islam et al. \cite{haq2009} used multiquadric radial basis functions in an RBF-based meshfree collocation method to numerically solve the RLW equation. Our method performed very well as we found that when simulating single solitons, we obtained a value of $L_2 \approx 2.07 \times 10^{-4}$ and a value of $L_\infty \approx 7.80 \times 10^{-5}$. When simulating the interaction of two solitons, we found that the conservation errors were less than 0.0014\% for $I_1$, 0.0005\% for $I_2$, and 0.0008\% for $I_3$. Also, we were able to simulate the evolution of undular bores up to time $t=250$ with accurate representation of the wave train formation.

Additional comparisons have been made to traditional finite difference (FDM) and finite element (FEM) methods. The compact finite difference schemes \cite{pan2015} achieve fourth-order spatial accuracy with $L_2$ errors of $O(10^{-5})$ for single-soliton propagation. Galerkin finite element methods \cite{mei2014,liu2013} demonstrate robust performance with conservation errors below 0.01\%. Spectral methods \cite{guo1988, kang2015} have shown exponential convergence rates; however, they require that the solution be smooth and that the geometry be relatively simple. In comparison, successful PINN approaches provided similar levels of accuracy as these other methods, while preserving the mesh-free characteristics of PINNs.

\section{Enhanced PINN Approaches}\label{sec:enhanced_pinns}

\subsection{Adaptive PINN Approach}\label{subsec:adaptive}
We propose an Adaptive PINN approach to adaptively balance between PDE residual and initial/boundary residual losses by employing trainable adaptive weighting factors for each residual component in our objective function.

The adaptive PINN loss function is formulated as
\begin{equation}
\mathcal{L}_{\text{adaptive}} = \sum_{j \in \{\text{PDE},\text{IC},\text{BC}\}} \left( \frac{1}{2} e^{-\lambda_j} \mathcal{L}_j + \frac{1}{2} \lambda_j \right)
\end{equation}
where, $\mathcal{L}_j$, denotes one of the three individual residual losses associated with PDE residual, initial condition residual and boundary condition residual, respectively. $\lambda_j$, represents a learnable parameter used to control the weights on the corresponding residual loss. The use of the exponential function $e^{-\lambda_j}$ ensures positive weights and provides smooth balancing among loss components. Both $\lambda_j$ and network weights $\theta$ are optimized simultaneously using gradient-based methods.

Both the loss and weights are optimized at the same time for an adaptive solution (the optimal saddle point) of the optimization problem. The gradients are calculated with respect to both $\theta$ and $\lambda_j$, as follows:

\begin{align}
\frac{\partial \mathcal{L}_{\text{adaptive}}}{\partial \theta} &= \sum_{j} \frac{1}{2} e^{-\lambda_j} \frac{\partial \mathcal{L}_j}{\partial \theta} \\
\frac{\partial \mathcal{L}_{\text{adaptive}}}{\partial \lambda_j} &= -\frac{1}{2} e^{-\lambda_j} \mathcal{L}_j + \frac{1}{2}
\end{align}
When the loss $\mathcal{L}_j$ is large, the gradient with respect to $\lambda_j$ will increase the weight of the corresponding term, thus pushing the optimization process toward those terms that were poorly satisfied by the physics constraint. The self-balancing nature of this mechanism helps to resolve issues of loss imbalance present in traditional PINNs, providing more robust convergence and improved accuracy.

\subsection{Conservative PINN Approach}\label{subsec:conservative}
We build on our previous work with the Adaptive approach to create a Conservative PINN approach which will force the enforcement of the three RLW conservation equations.

Our loss function will be modified so that it contains penalty terms for violation of each of the conserved quantities (mass, momentum, and energy). This will ensure that the final solution has been physically constrained by the fundamental laws governing the system.

Our Conservative PINN loss function is defined as follows:
\begin{equation}
\mathcal{L}_{\text{conservative}} = \mathcal{L}_{\text{adaptive}} + \lambda_{\text{cons}} \mathcal{L}_{\text{cons}},
\end{equation}
where $\mathcal{L}_{\text{adaptive}}$ is the adaptive loss function (Section~\ref{subsec:adaptive}), $\lambda_{\text{cons}}$ controls the influence of the conservation penalty, and $\mathcal{L}_{\text{cons}}$ is the conservation loss based on the three RLW conservation laws.

Each of the conserved quantities in our problem are calculated from the predicted solution as

\begin{align}
I_1(t) &= \int_{x_{\min}}^{x_{\max}} u(x,t;\theta)  dx \\
I_2(t) &= \int_{x_{\min}}^{x_{\max}} \left( u^2(x,t;\theta) + \mu \left( \frac{\partial u}{\partial x}(x,t;\theta) \right)^2 \right) dx \\
I_3(t) &= \int_{x_{\min}}^{x_{\max}} \left( u^3(x,t;\theta) + 3u^2(x,t;\theta) \right) dx
\end{align}
The conservation loss term then calculates the deviation of the above conservation laws from their initial values at $t=0$.
\begin{equation}
\mathcal{L}_{\text{cons}} = \sum_{k=1}^{3} \left( \frac{1}{N_t} \sum_{i=1}^{N_t} |I_k(t_i) - I_k(0)|^2 \right)
\end{equation}
In the above equation $N_t$ represents the number of time instances sampled in order to evaluate the conservation, and $t_i$ represent the individual time points sampled. The integrals above are evaluated numerically using the trapezoidal rule. The hyperparameter $\lambda_{\text{cons}}$ is used to control how strongly the conservation constraint is enforced. Typically, this value is between $10^{-4}$ to $10^{-5}$.

\subsection{Two-Stage Curriculum Training Strategy}\label{subsec:curriculum}

Our two-stage curriculum training strategy allows us to train our physics-informed neural networks (PINNs) to learn conservation-governed problems, like the interaction between two solitons. First, the PINN learns about the basic physics of the problem, and then it enforces the physical conservation laws. We use this method because if we enforce the conservation laws too early in the training process, they can severely limit the ability of the initial solution space and reduce the long-term accuracy and stability of the trained PINN.

We use a two-phase approach to carry out the two-stage curriculum training. Stage 1 trains the PINN using the PDE, ICs, and BCs, but does not enforce the conservation laws. The goal of stage 1 is to allow the neural network to find a solution manifold that includes the essential physics of the problem. The loss function used for the PINN during stage 1 is:

\begin{equation}
\mathcal{L}_{\text{Stage 1}}(\theta) = \mathcal{L}_{\text{PDE}}(\theta) + \mathcal{L}_{\text{IC}}(\theta) + \mathcal{L}_{\text{BC}}(\theta)
\end{equation}

where $\theta$ represents the neural network parameters. Self-adaptive weighting balances these components dynamically during training.

Once the PINN has converged during Stage~1, Stage~2 refines the PINN while enforcing the conservation laws. The loss function is extended with a penalty term for the deviation of the RLW conservation laws $I_1$, $I_2$, and $I_3$ (Eqs.~16--18):

\begin{equation}
\mathcal{L}_{\text{Stage 2}}(\theta) = \mathcal{L}_{\text{Stage 1}}(\theta) + \lambda_c \mathcal{L}_{\text{Conservation}}(\theta)
\end{equation}

where $\lambda_c$ is a tunable weighting coefficient that determines how much weight should be assigned to the conservation laws. The conservation loss function $\mathcal{L}_{\text{Conservation}}$ is calculated by computing the error in the conservation laws at random sampling points in time and comparing them to their values at the start of the simulation:
\begin{equation}
\mathcal{L}_{\text{Conservation}} = \mathbb{E}_{t \sim \mathcal{U}[0, T]} \left[ (I_1(t) - I_1(0))^2 + (I_2(t) - I_2(0))^2 + (I_3(t) - I_3(0))^2 \right]
\end{equation}

Both stages are performed in two phases. Stage~1 uses 50,000 Adam iterations and stage~2 uses 10,000 L-BFGS iterations with the conservation loss active ($\lambda_c = 10^{-5}$). For stage~2, we also lower the learning rate to avoid over-fitting of the conservation loss. By doing so, the PINN is allowed to learn a general solution before it is refined into a solution that adheres to the conservation laws.

\subsection{Causal Training Strategy}\label{subsec:causal}
Because of the problems with the integration over a long period of time (notably those related to the simulation of undular bores) we have adopted a strategy for the causality training that divides the temporal interval in successive temporal intervals such that it is possible to reduce the problem of the accumulation of errors during prolonged simulations.

The temporal domain $T = [0, T_{\text{final}}]$ can be divided into $N$ successive temporal sub-domains as follows:
\begin{equation}
T = \bigcup_{i=1}^{N} T_i, \quad \text{where} \quad T_i = [t_{i-1}, t_i].
\end{equation}
Each of these sub-domains has an extension $\Delta t = T_{\text{final}}/N$. In particular, in case of undular bore simulation ($T_{\text{final}} = 250$) we considered $N=5$, i.e., five successive sub-domains of 50-time unit each. The number $N=5$ represents a good balance between the need to reduce the computational cost and the need to ensure the continuity of the solution. A smaller value of $N$ implies a greater accumulation of errors, whereas a larger number of $N$ does not lead to a significantly greater accuracy, but increases the computational costs.

A separate PINN is trained sequentially on each sub-domains. The first sub-domain $T_1 = [0, 50]$ uses the analytical initial condition, while subsequent sub-domains $T_i$ ($i > 1$) use the predicted solution from the preceding sub-domain's final time step as the initial condition, ensuring temporal continuity.

Each PINN uses 8 hidden layers with 100 neurons per layer and the SiLU activation function. Training for each sub-domain consists of 20,000 Adam epochs (learning rate $10^{-3}$) followed by 5,000 L-BFGS iterations for fine-tuning. Self-adaptive loss balancing (Section~\ref{subsec:adaptive}) ensures robust convergence.

The final global solution $u_{\text{final}}(x,t)$ is constructed by stitching together the individual solutions from each temporal sub-domain:
\begin{equation}
u_{\text{final}}(x,t) = 
\begin{cases} 
u_{\theta_1}(x,t) & \text{if } t \in [t_0, t_1] \\
u_{\theta_2}(x,t) & \text{if } t \in (t_1, t_2] \\
\vdots \\
u_{\theta_N}(x,t) & \text{if } t \in (t_{N-1}, t_N]
\end{cases}
\end{equation}
where $u_{\theta_i}(x,t)$ represents the solution from the PINN trained on the $i$-th temporal sub-domain.

\subsection{Integration of Approaches for Different Test Cases}\label{subsec:integration}
The enhanced PINN approaches are used strategically depending upon the characteristics of each problem. In terms of single solitons and two-soliton, both PINNs (Adaptive and Conservative) were trained in the full temporal domain but due to the interaction between the two solitons and a possible loss of stability during the training process, the conservative PINN was trained in two stages using the curriculum described in Section~\ref{subsec:curriculum}. The Adaptive PINN and the Conservative PINN both use causal training as described in Section~\ref{subsec:causal} when training the PINN on undular bores, since the errors accumulate throughout time, and therefore would otherwise be difficult to train accurately.

\section{Experimental Setup}\label{sec:experimental_setup}

The computational configurations of all PINNs are detailed in this section. A base model configuration is defined; we modify each test case’s base model configuration to meet its own unique computational issues.

\subsection{Base Configuration}\label{subsec:base_config}

All models share the following core configuration:

\begin{itemize}
    \item \textbf{Architecture:} Feedforward Neural Network (FNN)
    \item \textbf{Activation function:} SiLU
    \item \textbf{Optimization:} Two-stage optimization methodology, first using Adam (Learning rate $= 10^{-3}$), then using L-BFGS for fine tuning.
    \item \textbf{Loss function:} Self-adaptive formulation
    \[
    \mathcal{L}_{\text{adaptive}} = \sum_{i \in \{\text{PDE}, \text{IC}, \text{BC}\}} 
    \left( \tfrac{1}{2} e^{-\lambda_i} \mathcal{L}_i + \tfrac{1}{2} \lambda_i \right)
    \]
\end{itemize}

Uniformly distributed spatial and temporal collocation points across their respective domains were used for sampling loss functions associated with the partial differential equation (PDE), initial conditions and boundary conditions. For initializing network weights, Kaiming Uniform was employed. The conservation hyperparameter $\lambda_{\text{cons}}$ was determined empirically as a trade-off between enforcing conservation and maintaining training stability.

Table~\ref{tab:config} summarizes test-case-specific parameter variations.

\begin{table}[h!]
\centering
\small
\caption{Model configurations across test cases.}
\label{tab:config}
\begin{tabular}{lccc}
\toprule
\textbf{Parameter} & \textbf{Single Soliton} & \textbf{Two-Soliton} & \textbf{Undular Bore} \\
\midrule
Network layers $\times$ width & 8 $\times$ 50 & 8 $\times$ 100 & 8 $\times$ 100 \\
Collocation points (PDE / IC / BC) & 20k / 5k / 5k & 40k / 10k / 10k & 40k / 10k / 10k \\
Adam epochs & 30,000 & 50,000 & 20,000$^{\dagger}$ \\
L-BFGS iterations & 5,000 & 10,000 & 5,000$^{\dagger}$ \\
Conservation weight $\lambda_{\text{cons}}$ & $10^{-4}$ & $10^{-5}$ & $10^{-5}$ \\
Special strategy & --- & Two-stage$^{\ddagger}$ & Causal ($N=5$) \\
\bottomrule
\end{tabular}

\vspace{2mm}
\raggedright
\footnotesize
$^{\dagger}$ Per temporal window. \\
$^{\ddagger}$ Stage~1: 50k Adam, Stage~2: 10k L-BFGS with conservation.
\end{table}

\subsection{Performance Metrics and Environment}\label{subsec:metrics}

The solution accuracy was evaluated via the relative $L_2$ and $L_\infty$ error norms for:
\[
L_2 = \frac{\|u_{\text{PINN}} - u_{\text{exact}}\|_2}{\|u_{\text{exact}}\|_2}, 
\qquad 
L_\infty = \frac{\max |u_{\text{PINN}} - u_{\text{exact}}|}{\max |u_{\text{exact}}|}.
\]

Conservation of physical properties was assessed by calculating numerical integrals of the PINN solution to determine how well each conserved property was preserved in time. The conservation errors were determined by the equation:
\[
\text{Error}_k = \frac{|I_k(t) - I_k(0)|}{|I_k(0)|} \times 100\%.
\]

where $k$ represents one of three conserved quantities (mass, momentum, or total energy). All calculations were performed on a High Performance Computing Node containing a single NVIDIA P100 Graphics Processing Unit (GPU).

\section{Results and Discussion}\label{sec:results}

The Adaptive and Conservative PINN approaches are assessed in detail over three benchmark problems for the RLW model. The relative $L_2$ and $L_\infty$ errors are used as the two main accuracy metrics for our assessment and compared with established numerical models that provide high accuracy. Additionally, we assess conservation properties to determine the degree of physical fidelity of the solutions obtained from these models. Through this examination, we find that the efficacy of each approach depends on the specific problem being modeled, and that there are important tradeoffs between loss adaptation and enforcing the conservation laws explicitly.

\subsection{Single Soliton Propagation}\label{subsec:single_soliton}
\begin{figure}[ht!]
    \centering
    \includegraphics[width=0.8\textwidth]{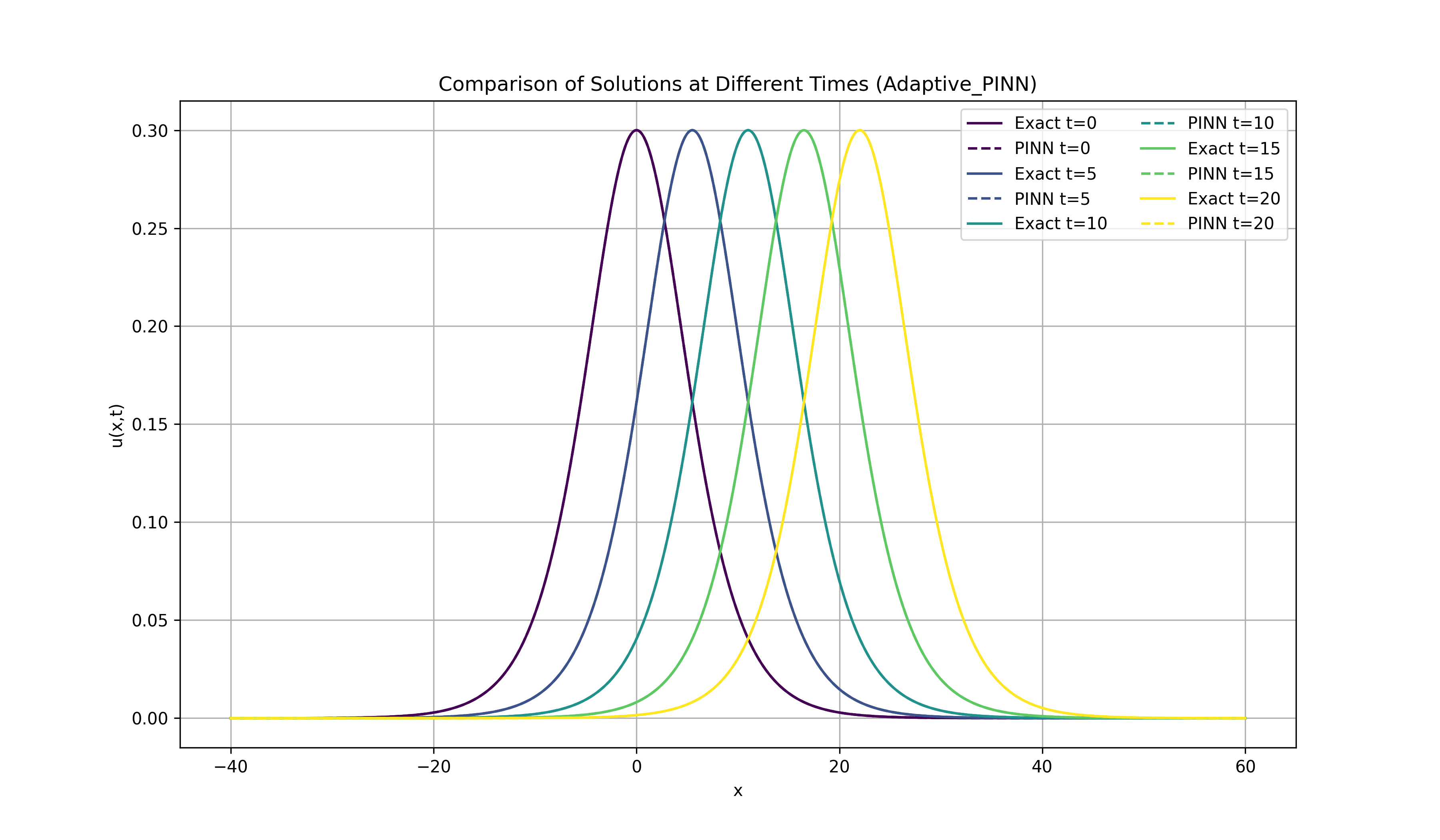}
    \caption{Wave profile comparison at different times for the Adaptive PINN solution versus the exact analytical solution for the single soliton case.}
    \label{fig:adaptive_pinn_profile}
\end{figure}

\begin{figure}[ht!]
    \centering
    \includegraphics[width=0.8\textwidth]{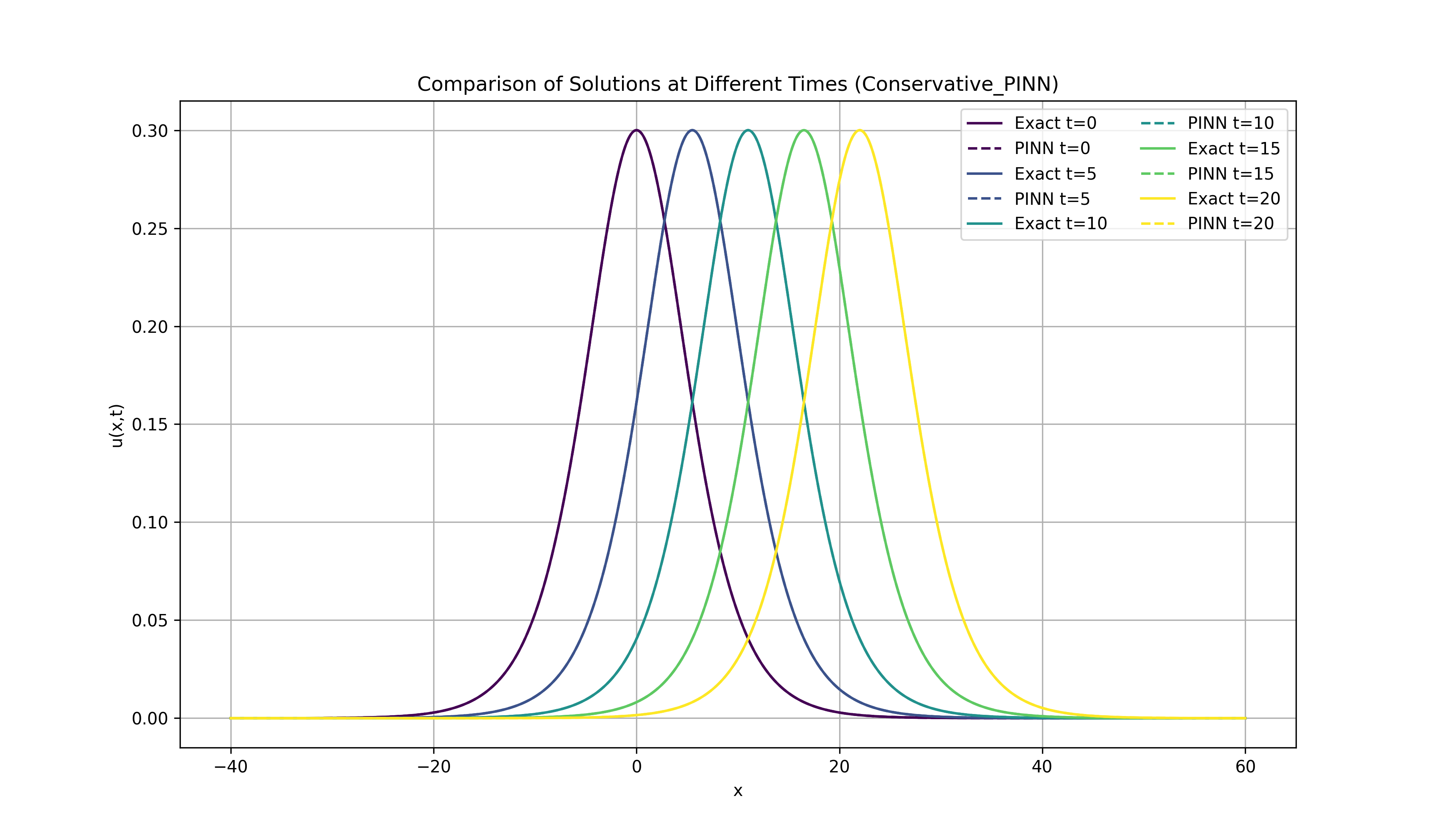}
    \caption{Wave profile comparison at different times for the Conservative PINN solution versus the exact analytical solution.}
    \label{fig:conservative_pinn_profile}
\end{figure}

\begin{figure}[ht!]
    \centering
    \includegraphics[width=0.75\textwidth]{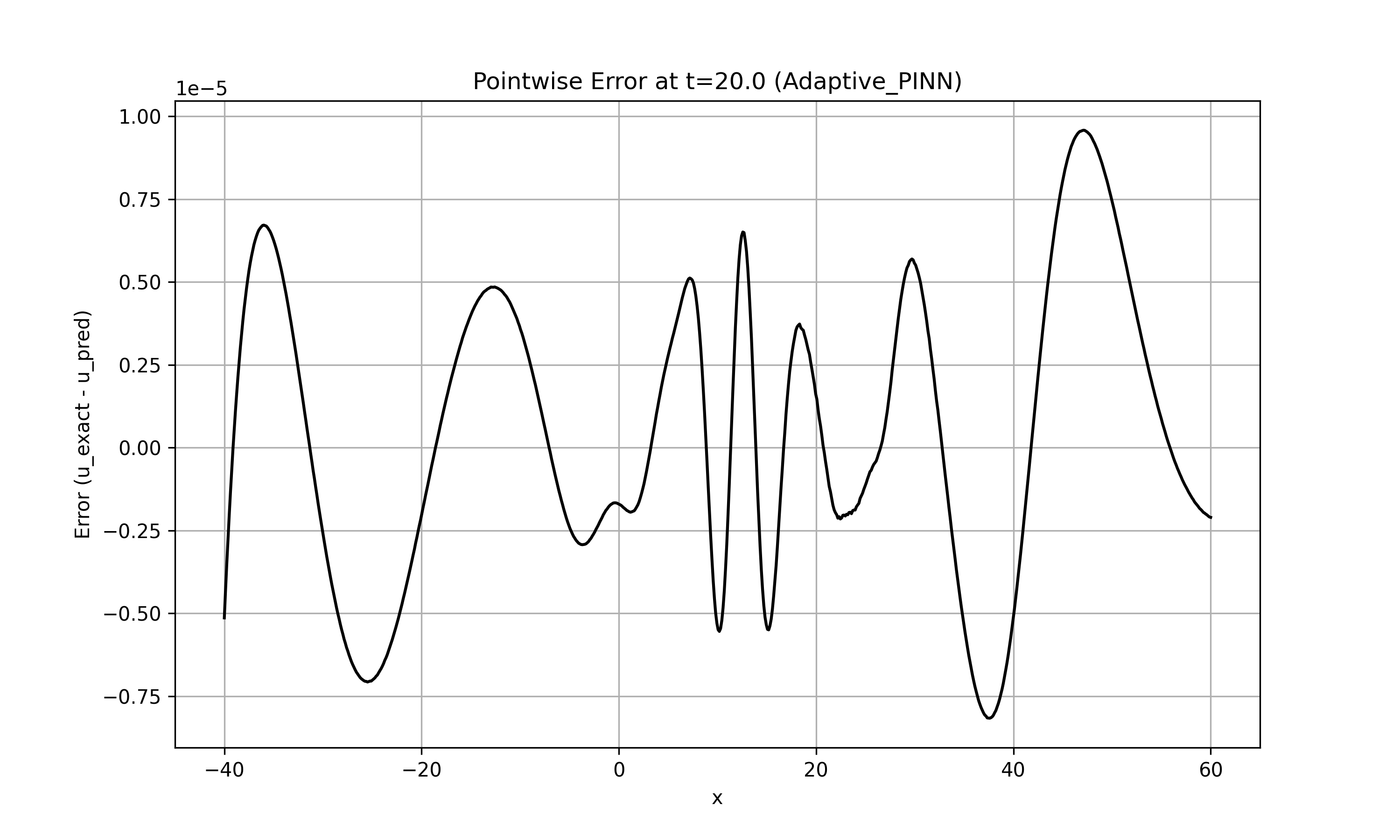}
    \hfill
    \includegraphics[width=0.75\textwidth]{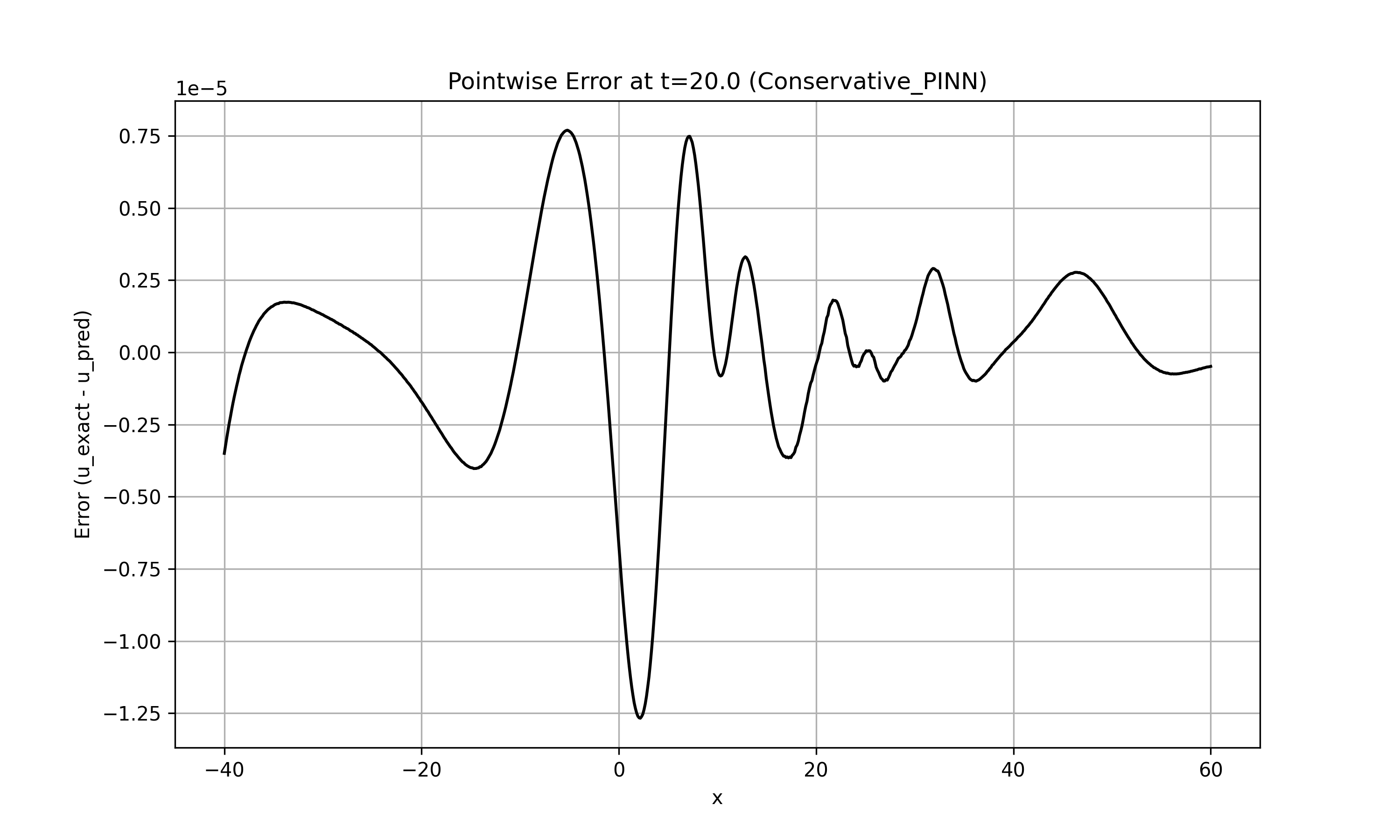}
    \caption{Pointwise absolute error at $t=20$ for (a) Adaptive PINN and (b) Conservative PINN.}
    \label{fig:pointwise_error}
\end{figure}

\begin{figure}[ht!]
    \centering
    \includegraphics[width=0.8\textwidth]{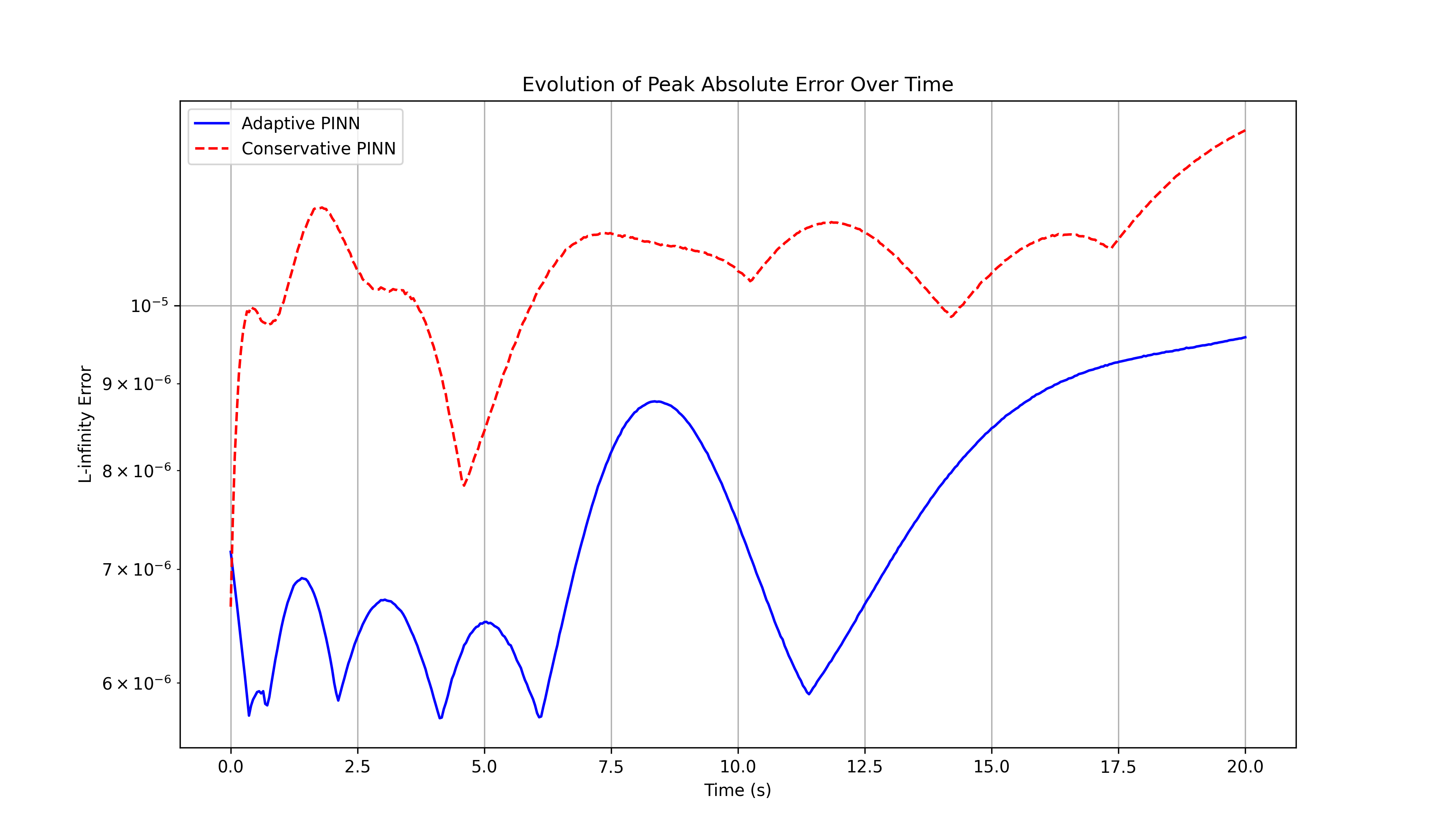}
    \caption{Evolution of the peak absolute error ($L_{\infty}$) over time for the Adaptive and Conservative PINN models.}
    \label{fig:peak_error}
\end{figure}

Table~\ref{tab:single_soliton_results} summarizes the final errors and conserved conservation laws at $t = 20$. Wave profiles at different times are shown in Figures~\ref{fig:adaptive_pinn_profile} and \ref{fig:conservative_pinn_profile}, pointwise absolute errors in Figure~\ref{fig:pointwise_error}, and $L_\infty$ error evolution in Figure~\ref{fig:peak_error}.

\begin{table}[htbp]
\centering
\caption{Final error and conservation law comparison for the single soliton test case at $t=20$.}
\label{tab:single_soliton_results}
\begin{tabular}{@{}lccccc@{}}
\toprule
\textbf{Method} & \textbf{$L_\infty$ Error} & \textbf{$L_2$ Error} & \textbf{$I_1$} & \textbf{$I_2$} & \textbf{$I_3$} \\
\midrule
Haq \& Ali \cite{haq2009} & $1.32 \times 10^{-5}$ & $4.32 \times 10^{-5}$ & 3.980 & 0.810 & 2.579 \\
Adaptive PINN & $9.58 \times 10^{-6}$ & $4.36 \times 10^{-5}$ & 3.980 & 0.810 & 2.579 \\
Conservative PINN & $1.27 \times 10^{-5}$ & $3.37 \times 10^{-5}$ & 3.980 & 0.810 & 2.579 \\
\bottomrule
\end{tabular}
\end{table}

Both enhanced PINN approaches achieve substantial accuracy improvements, with errors of $O(10^{-5})$ comparable to the established high-accuracy benchmark \cite{haq2009}.

The Adaptive PINN achieves the lowest pointwise error ($L_\infty = 9.58 \times 10^{-6}$, Figure~\ref{fig:peak_error}), accurately capturing the soliton's shape, amplitude, and velocity with minimal deviation from the exact solution (Figure~\ref{fig:adaptive_pinn_profile}).

\begin{figure}[ht!]
    \centering
    \includegraphics[width=0.8\textwidth]{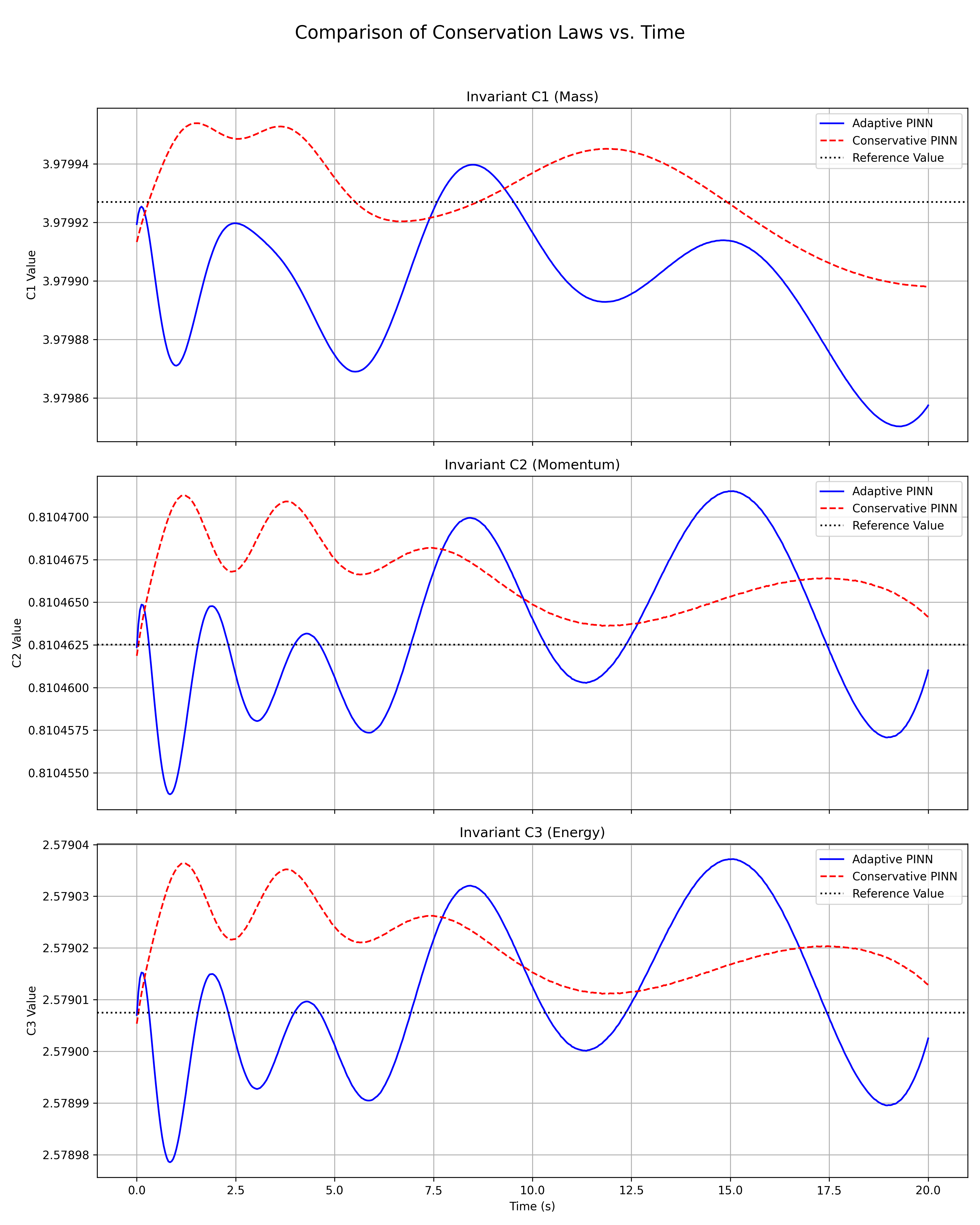}
    \caption{Temporal evolution of the conservation laws ($I_1$: Mass, $I_2$: Momentum, $I_3$: Energy) for both enhanced PINN approaches, compared to the constant reference value.}
    \label{fig:invariants_comparison}
\end{figure}

The Conservative PINN shows the lowest total error ($L_2 = 3.37 \times 10^{-5}$) and the best conservation of the physical laws. The figure~\ref{fig:invariants_comparison} illustrates that although the Adaptive PINN presents slight deviations in mass ($I_1$) and momentum ($I_2$), the Conservative PINN conserves the quantity of $I_2$ and $I_3$, with little deviation from its original value. These preserved physical laws provide a strong constraint for inducing stability and accuracy for obtaining the best results in terms of total $L_2$ error.

Conservation laws can provide very useful inductive bias in favor of stable, structure-preserving solutions such as a single soliton. Although both methods have been shown to be highly accurate, they excel in different ways; the Adaptive PINN has excellent pointwise accuracy and the Conservative PINN has excellent overall fidelity and conservation of physical laws.

\subsection{Two-Soliton Interaction}\label{subsec:two_soliton}
\begin{figure}[htbp]
    \centering
    \includegraphics[width=0.8\textwidth]{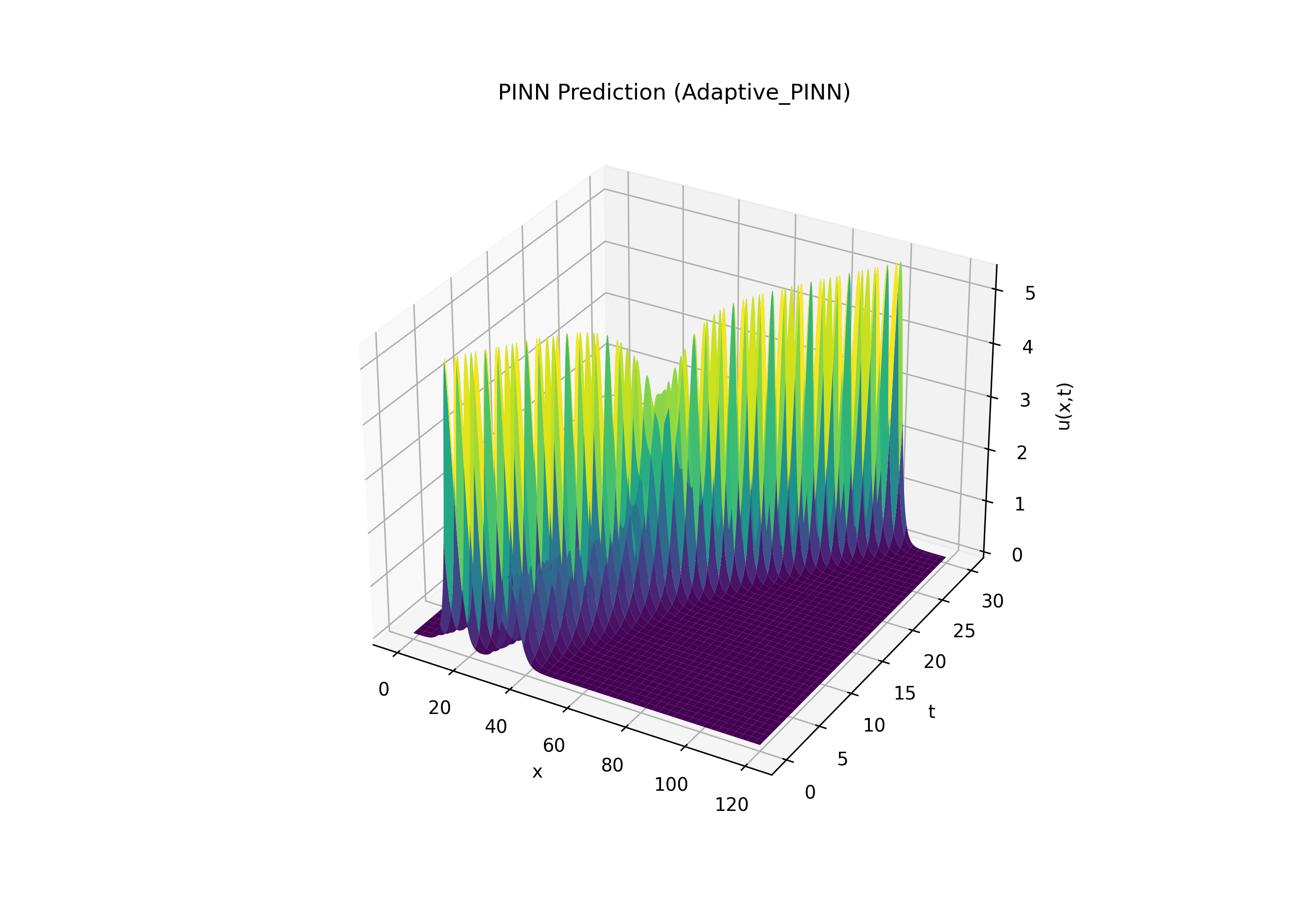}
    \caption{Space-time surface plot of the two-soliton interaction predicted by the Adaptive PINN.}
    \label{fig:adaptive_pinn_surface}
\end{figure}

\begin{figure}[htbp]
    \centering
    \includegraphics[width=0.8\textwidth]{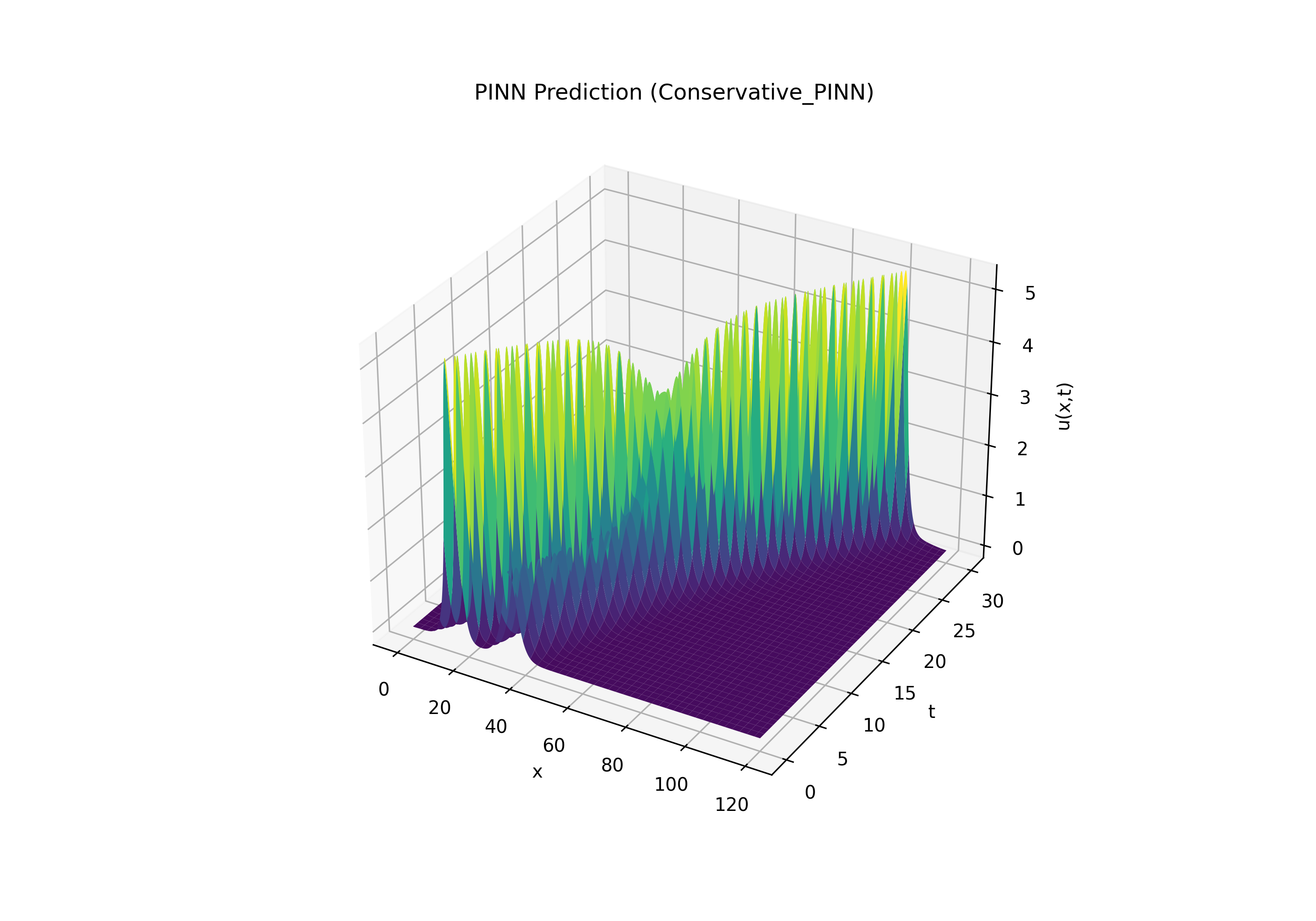}
    \caption{Space-time surface plot of the two-soliton interaction predicted by the Conservative PINN, showing spurious oscillations post-collision.}
    \label{fig:conservative_pinn_surface}
\end{figure}

\begin{figure}[htbp]
    \centering
    \includegraphics[width=\textwidth]{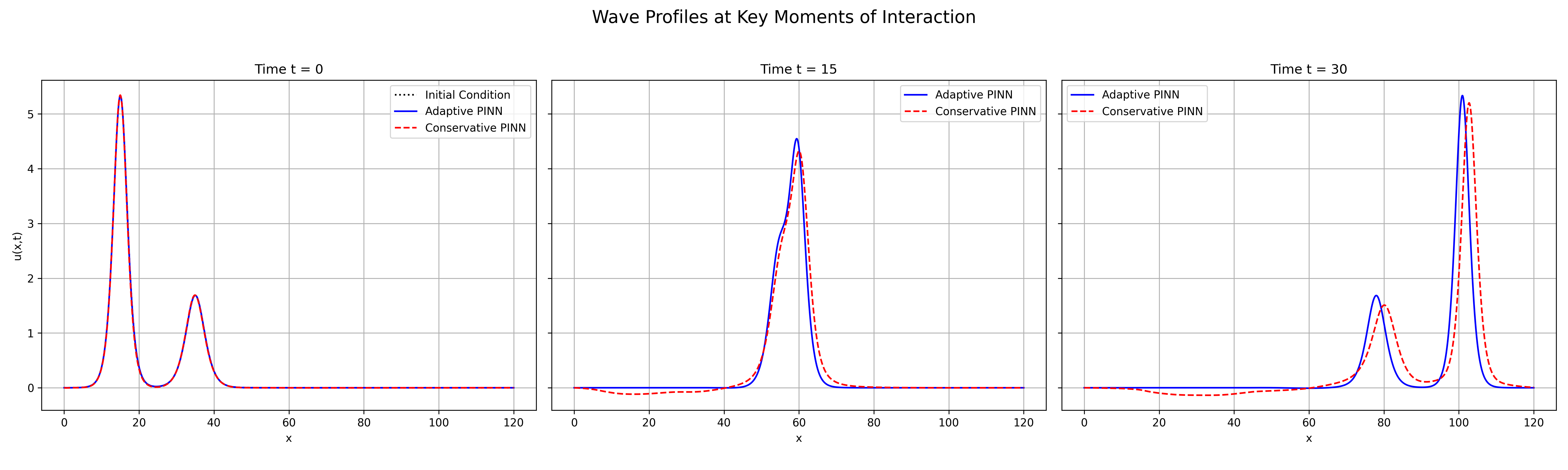}
    \caption{Snapshot comparison of wave profiles at the initial ($t=0$), interaction ($t=15$), and final ($t=30$) times for both methods.}
    \label{fig:key_moments}
\end{figure}

\begin{figure}[htbp]
    \centering
    \includegraphics[width=\textwidth]{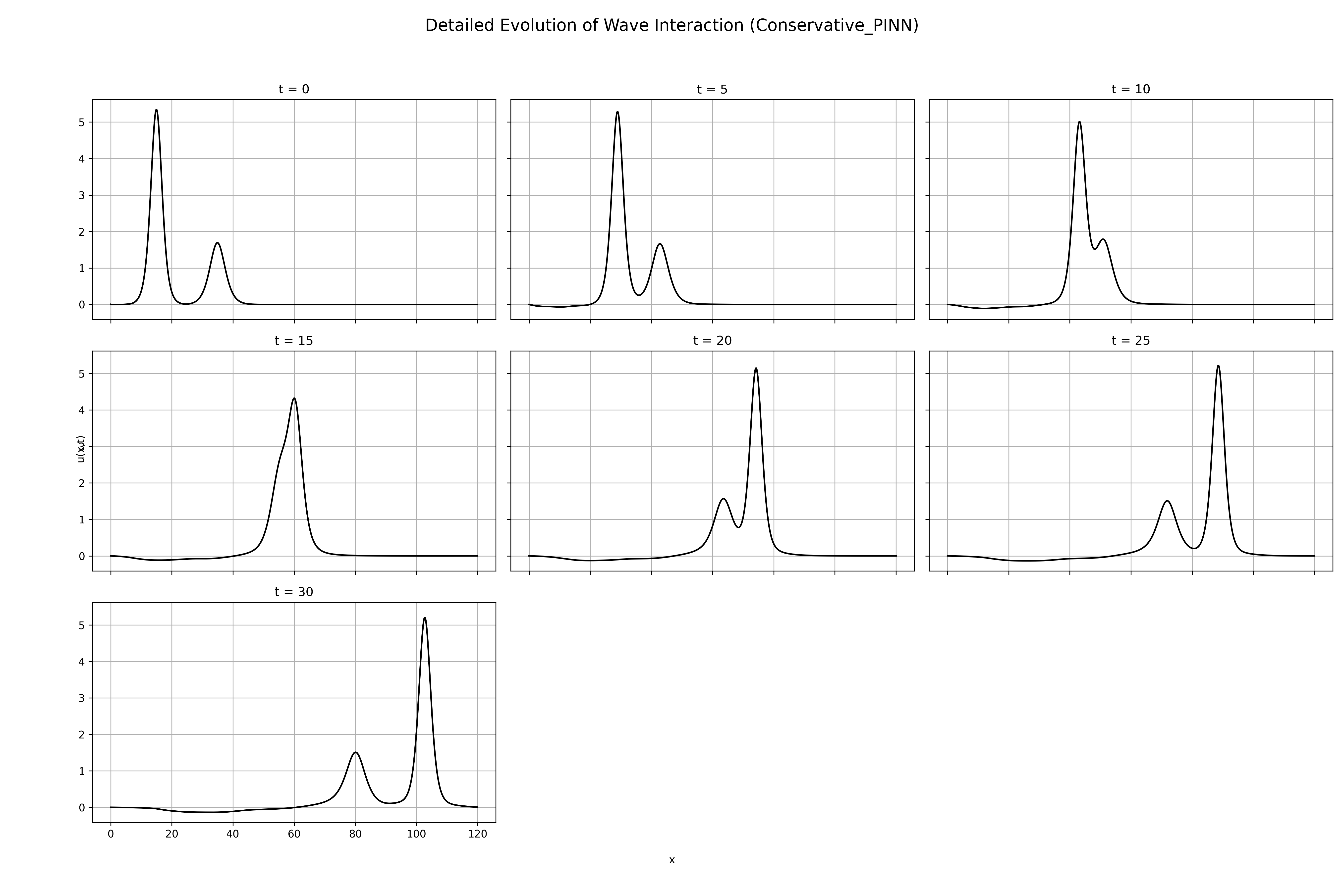}
    \caption{Detailed evolution of the wave interaction for the Conservative PINN, highlighting the development of instabilities during the collision.}
    \label{fig:conservative_detailed}
\end{figure}

\begin{figure}[htbp]
    \centering
    \includegraphics[width=0.8\textwidth]{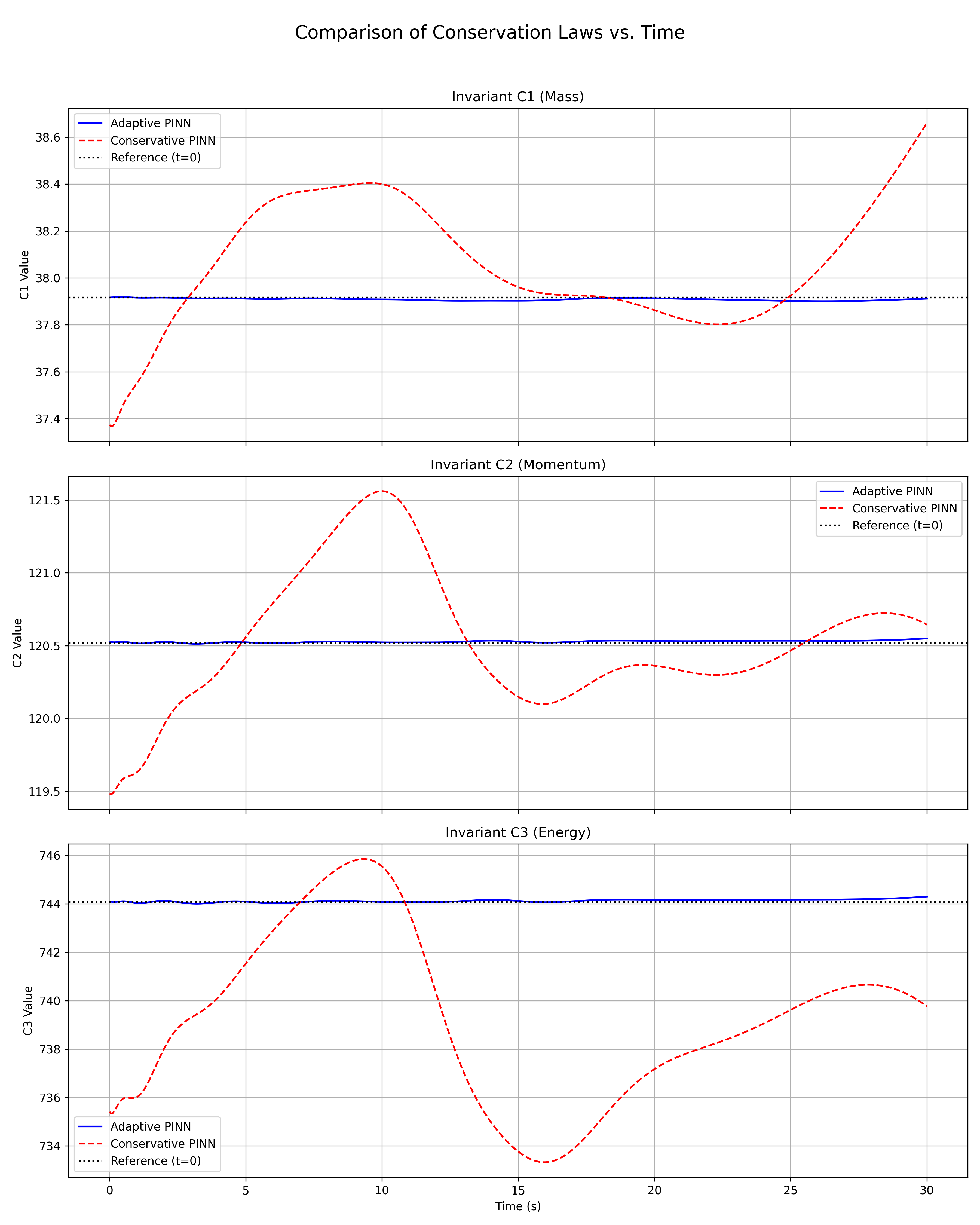}
    \caption{Temporal evolution of the conservation laws ($I_1$: Mass, $I_2$: Momentum, $I_3$: Energy) for both enhanced PINN approaches. Note the Conservative PINN's incorrect initial value for $I_3$.}
    \label{fig:conserved_evolution}
\end{figure}

Interaction between two solitons has been identified as one of the most difficult problems to solve due to the high level of non-linearity experienced during collisions. The success of solving this problem can be quantified by ensuring that the solitons experience an elastic collision resulting in solitons emerging from the collision with their original amplitude and shape. Soliton properties are shown in tables~\ref{tab:soliton_properties}, and \ref{tab:conservation_properties}, show the properties associated with conserved quantities. Figures~\ref{fig:adaptive_pinn_surface}--\ref{fig:conserved_evolution} provide visualization of the space-time evolution of the collision, key dynamics of the collision, and conserved quantities.

\begin{table}[htbp]
\centering
\small
\caption{Soliton properties before and after interaction.}
\label{tab:soliton_properties}
\begin{tabular}{lcccccc}
\toprule
\textbf{Method} & \textbf{Time} & \textbf{Wave 1 Amp.} & \textbf{Wave 1 Pos.} & \textbf{Wave 2 Amp.} & \textbf{Wave 2 Pos.} \\
\midrule
Haq \& Ali \cite{haq2009} & 0 & 5.333 & 15.00 & 1.688 & 35.00 \\
Haq \& Ali \cite{haq2009} & 30 & 5.331 & 100.7 & 1.683 & 77.90 \\
Adaptive PINN & 0 & 5.333 & 15.00 & 1.687 & 34.98 \\
Adaptive PINN & 30 & 5.334 & 100.98 & 1.684 & 77.94 \\
Conservative PINN & 0 & 5.342 & 15.00 & 1.691 & 34.92 \\
Conservative PINN & 30 & 5.201 & 102.72 & 1.509 & 80.10 \\
\bottomrule
\end{tabular}
\end{table}

\begin{table}[htbp]
\centering
\caption{Conservation law comparison at key time instants for two-soliton interaction.}
\label{tab:conservation_properties}
\begin{tabular}{lcccc}
\toprule
\textbf{Method} & \textbf{Time} & $\mathbf{I_1}$ \textbf{(Mass)} & $\mathbf{I_2}$ \textbf{(Momentum)} & $\mathbf{I_3}$ \textbf{(Energy)} \\
\midrule
Adaptive PINN & 0 & 37.92 & 120.52 & 744.08 \\
Conservative PINN & 0 & 37.37 & 119.48 & 735.41 \\
Adaptive PINN & 15 & 37.90 & 120.53 & 744.12 \\
Conservative PINN & 15 & 37.96 & 120.15 & 733.77 \\
Adaptive PINN & 30 & 37.91 & 120.55 & 744.30 \\
Conservative PINN & 30 & 38.66 & 120.64 & 739.77 \\
\bottomrule
\end{tabular}
\end{table}

We clearly see that this test case is a reversal of what we observed in the single soliton case; it confirms that our PINNs are going to be problem independent. While the Conservative PINN was much better than the Adaptive approach for a single soliton, it clearly does poorly when there are strong non-linear interactions between solitons.

The Conservative PINN is also unable to conserve the elastic collision property. Table~\ref{tab:soliton_properties} presents evidence of the larger soliton's amplitude decay from 5.34 to 5.20 (2.6\% error) and the smaller soliton's amplitude decay from 1.69 to 1.51 (10.7\% error). We also see large phase errors in the soliton locations at t=30. Figures~\ref{fig:conservative_pinn_surface} and~\ref{fig:conservative_detailed} illustrate why this is happening: they present spurious oscillations both before and after the collision. The model cannot accurately resolve the soliton interaction morphologies, and its conservation constraints prevent recovery.

The Adaptive PINN has an exceptionally high level of performance in contrast to that shown by the Conservative PINN. This is shown in Table~\ref{tab:soliton_properties}, where there is nearly perfect preservation of the amplitude of the waves (Wave~1: 5.333 $\rightarrow$ 5.334) and high degree of positional accuracy with all positional errors below 1\%. Figures~\ref{fig:adaptive_pinn_surface} and~\ref{fig:key_moments} shows the solitons have emerged with their original shapes and velocities unaltered as the result of their passage through the computational domain.

This performance difference between the two approaches is clearly illustrated by the results presented in Table~\ref{tab:conservation_properties} and Figure~\ref{fig:conserved_evolution}. The Conservative PINN shows smaller relative deviations from its own initial values for $I_2$ and $I_3$, but these initial values are incorrect ($I_3 = 735.41$ vs.~744.08 for the Adaptive case), indicating a poor initial approximation. The Adaptive PINN consistently provides accurate representations of the values of $I_2$ and $I_3$, while maintaining values for $I_1$ that are consistent with the physics of the problem, albeit with some deviation.

The success of this result has demonstrated a key insight; that the benefits of enforcing conservation law enforcement depend on the type of problem being addressed. In problems with highly dynamic interaction, then there will be an increased need for adaptive flexibility. Enforcing strict adherence to global conservation laws may hinder optimization by overly constraining the network in addition to creating a difficult optimization problem (i.e., complicated optimization landscape). The self-balancing mechanism in the Adaptive PINN allows it to give priority to the loss of PDE residual during collision, giving sufficient freedom to learn complex nonlinear interactions without introducing constraint that would be counter-productive.

\subsection{Undular Bore Development}\label{subsec:undular_bore}
The undular bore test evaluates long-time integration capabilities and stability. We consider both a gentle slope ($d=5$) and a steep slope ($d=2$) case with a final time of $t=250$, simulated using the causal training strategy. Performance is assessed by the ability to accurately predict the position and amplitude of the leading undulations in the developed wave train.

\subsubsection{Gentle Slope Case (d=5)}\label{subsubsec:gentle_slope}
\begin{figure}[htbp]
\centering
    \includegraphics[width=0.8\textwidth]{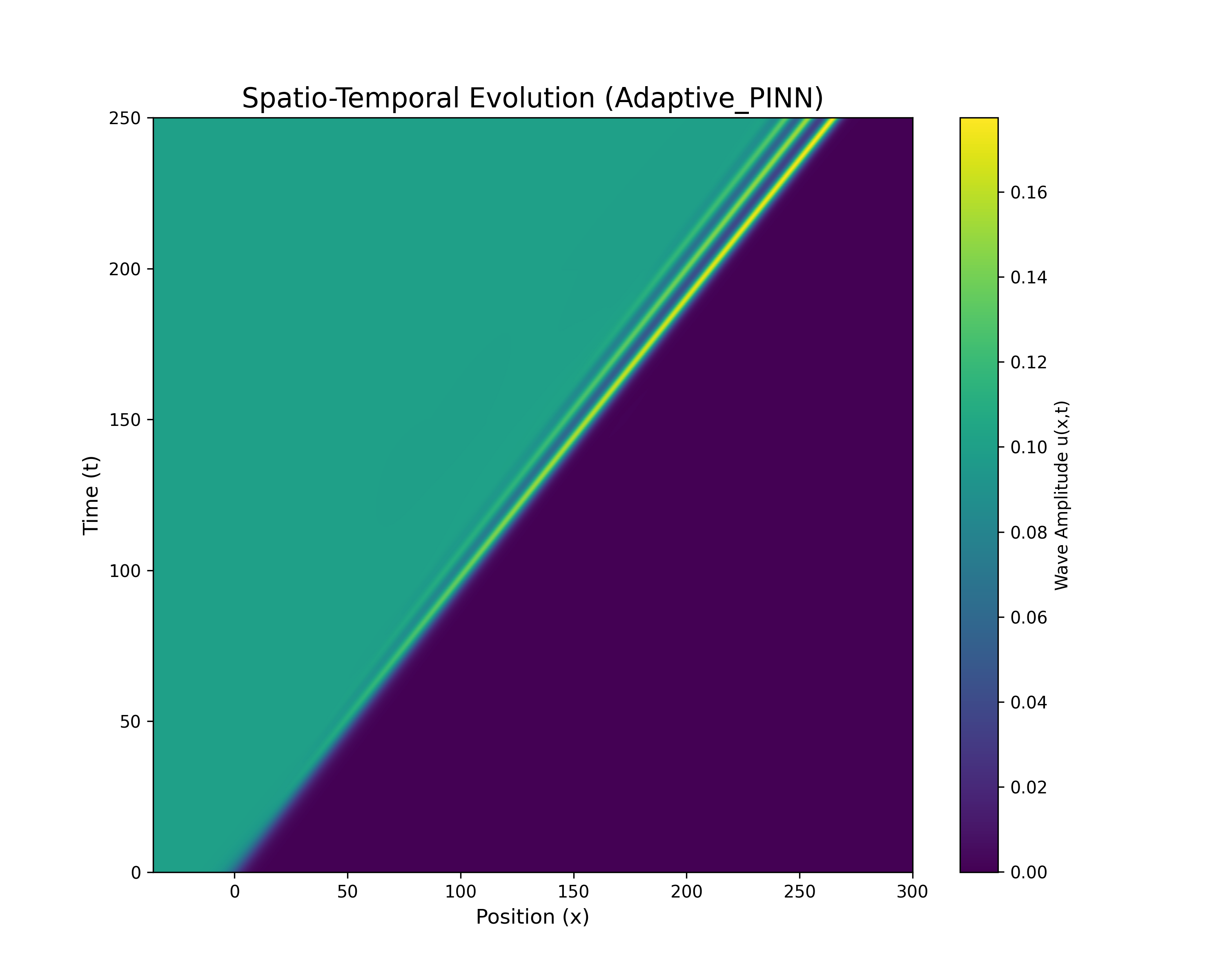}
    \caption{Spatio-temporal heatmap of the undular bore evolution ($d=5$) predicted by the Adaptive PINN.}
    \label{fig:adaptive_pinn_gentle}
\end{figure}

\begin{figure}[htbp]
\centering
    \includegraphics[width=0.8\textwidth]{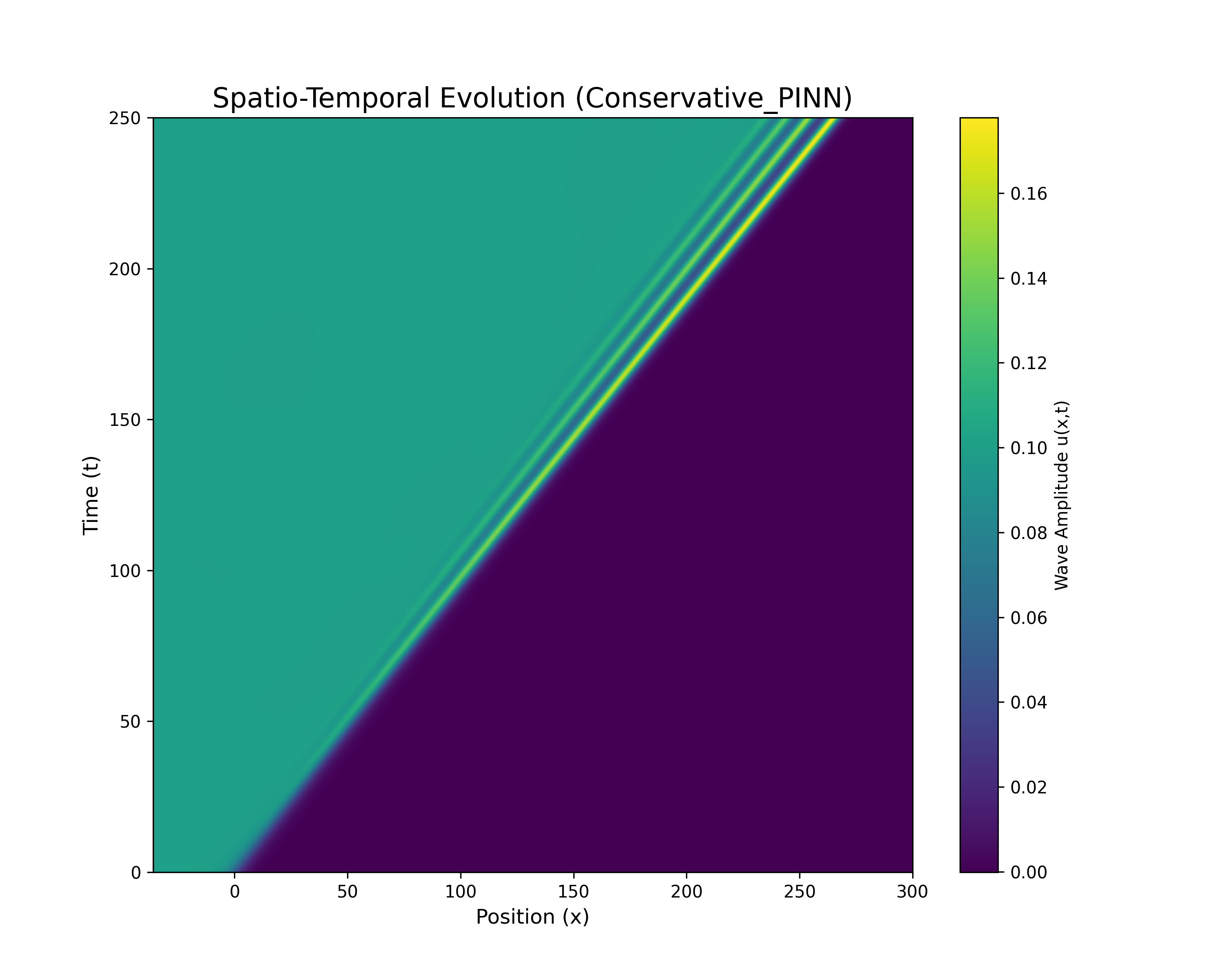}
    \caption{Spatio-temporal heatmap of the undular bore evolution ($d=5$) predicted by the Conservative PINN.}
    \label{fig:conservative_pinn_gentle}
\end{figure}

\begin{figure}[htbp]
\centering
    \includegraphics[width=0.8\textwidth]{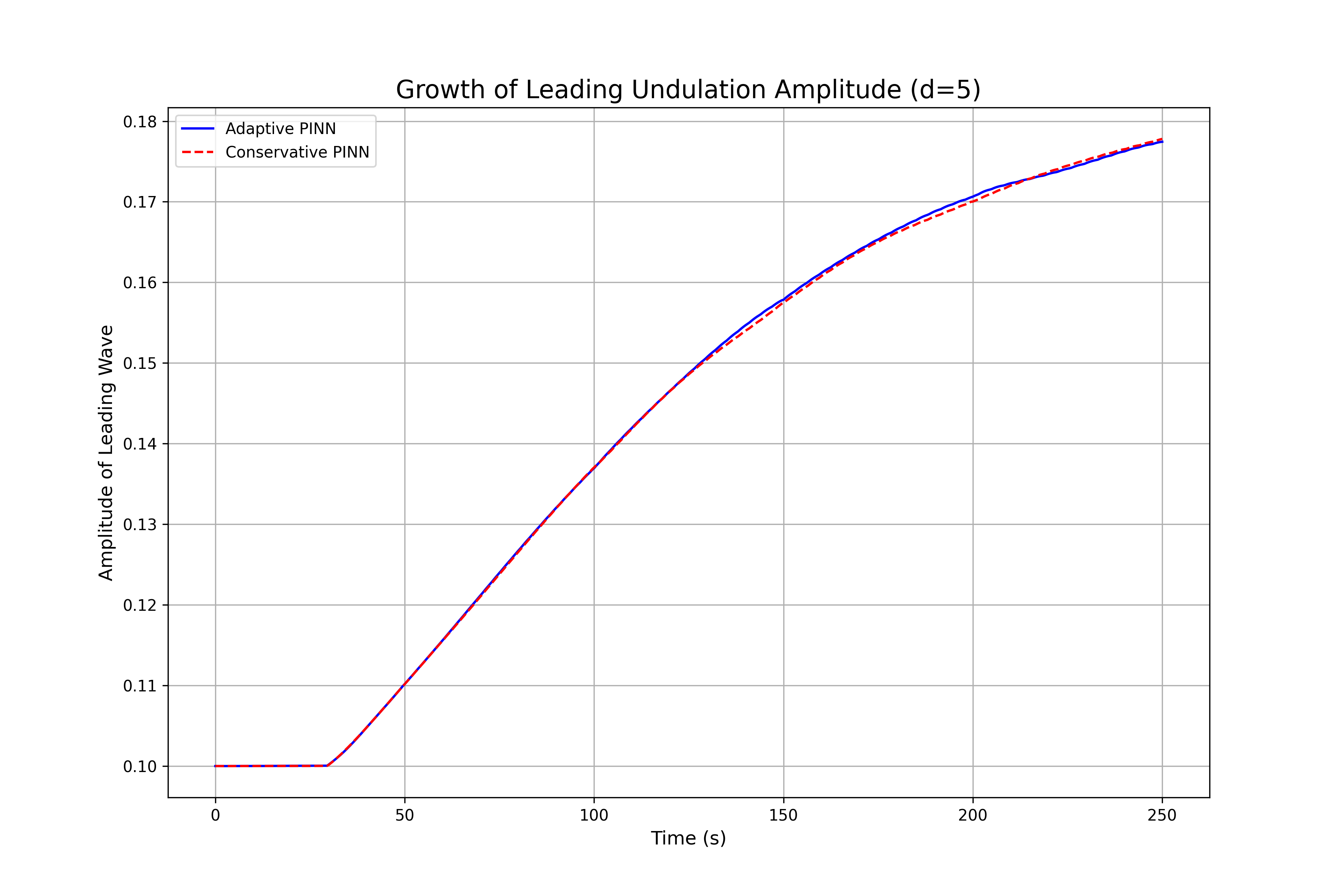}
    \caption{Growth of the leading undulation amplitude (top) and its position over time (bottom) for the $d=5$ case.}
    \label{fig:leading_wave_gentle}
\end{figure}

Results for the gentle slope case are summarized in Tables~\ref{tab:gentle_slope_results} and \ref{tab:gentle_slope_invariants}, with spatio-temporal evolution shown in Figures~\ref{fig:adaptive_pinn_gentle}--\ref{fig:conservative_pinn_gentle} and leading wave growth in Figure~\ref{fig:leading_wave_gentle}.

\begin{table}[htbp]
\centering
\caption{Final undulation properties at $t=250$ for the gentle slope ($d=5$) case.}
\label{tab:gentle_slope_results}
\begin{tabular}{lccc}
\toprule
\textbf{Method} & \textbf{Undulation} & \textbf{Position} & \textbf{Amplitude} \\
\midrule
Haq \& Ali \cite{haq2009} & Leading & 245.0 & 0.1779 \\
Adaptive PINN & Wave 1 & 265.1 & 0.1774 \\
Conservative PINN & Wave 1 & 265.1 & 0.1778 \\
Haq \& Ali \cite{haq2009} & Second & 253.9 & 0.1534 \\
Adaptive PINN & Wave 2 & 253.9 & 0.1527 \\
Conservative PINN & Wave 2 & 254.0 & 0.1527 \\
\bottomrule
\end{tabular}
\end{table}

\begin{table}[htbp]
\centering
\caption{Leading wave amplitude at key times for the $d=5$ case.}
\label{tab:gentle_slope_invariants}
\begin{tabular}{ccc}
\toprule
\textbf{Time} & \textbf{Adaptive PINN} & \textbf{Conservative PINN} \\
\midrule
0 & 0.1000 & 0.1000 \\
50 & 0.1102 & 0.1102 \\
100 & 0.1370 & 0.1370 \\
150 & 0.1578 & 0.1575 \\
200 & 0.1706 & 0.1700 \\
250 & 0.1774 & 0.1778 \\
\bottomrule
\end{tabular}
\end{table}

\textbf{Discussion for d=5:} For gentle initial conditions, both approaches demonstrate nearly indistinguishable accuracy. Tables~\ref{tab:gentle_slope_results} and \ref{tab:gentle_slope_invariants} and Figure~\ref{fig:leading_wave_gentle} show virtually identical amplitude growth, extremely close to the reference values. The Conservative PINN's final amplitude (0.1778) is marginally closer to the reference (0.1779) than the Adaptive PINN's (0.1774).

\begin{figure}[htbp]
\centering
    \includegraphics[width=0.8\textwidth]{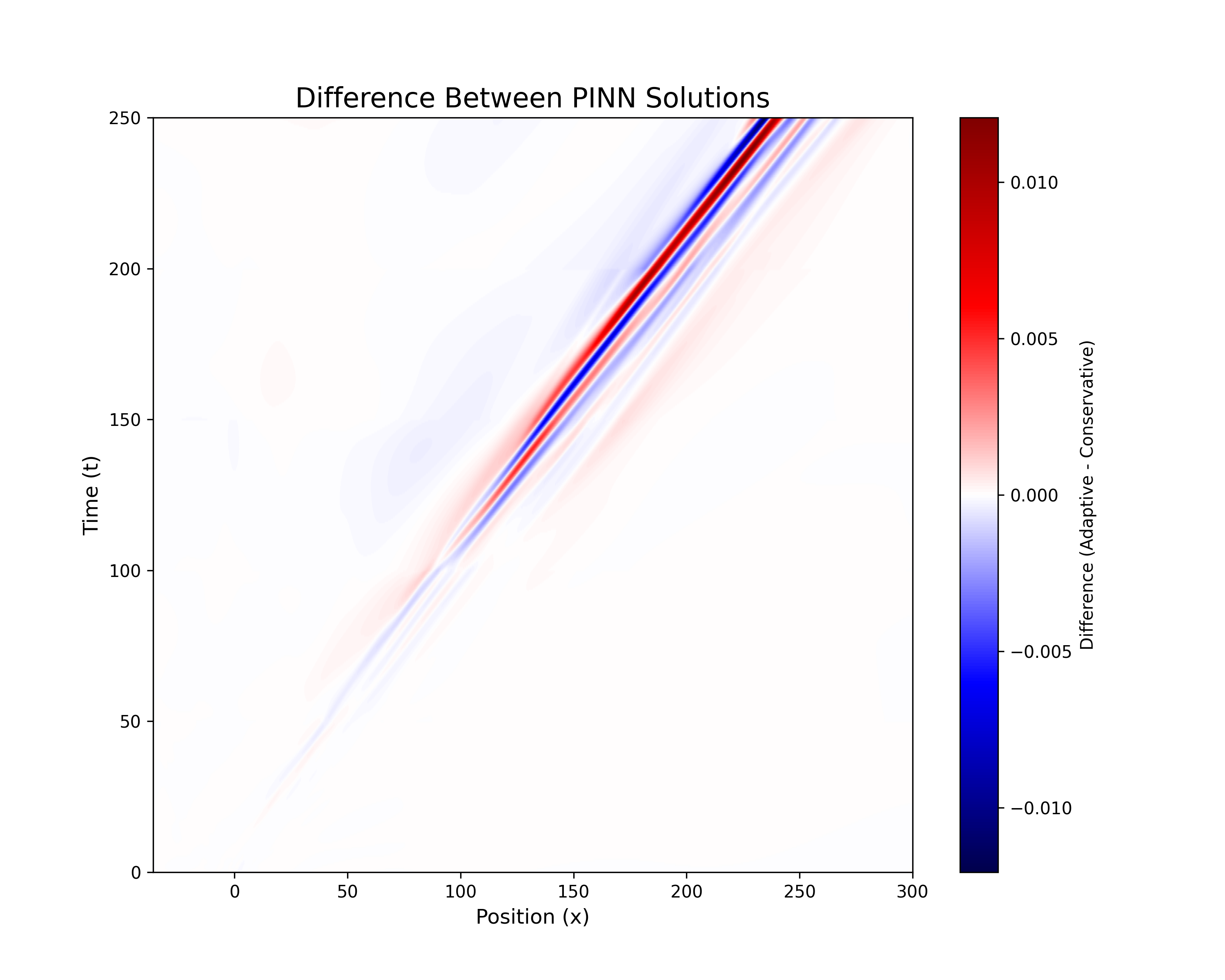}
    \caption{Difference plot (Adaptive - Conservative) of the solutions for the $d=5$ case, highlighting the minimal discrepancies.}
    \label{fig:difference_heatmap}
\end{figure}

Table~\ref{tab:gentle_slope_results} shows that both methods successfully capture undular bore evolution. The difference heatmap in Figure~\ref{fig:difference_heatmap} confirms minimal solution discrepancy ($O(10^{-3})$), orders of magnitude smaller than the wave amplitudes. For less challenging long-time evolution, both approaches are equally effective.

\subsubsection{Steep Slope Case (d=2)}\label{subsubsec:steep_slope}
\begin{figure}[htbp]
\centering
    \includegraphics[width=0.8\textwidth]{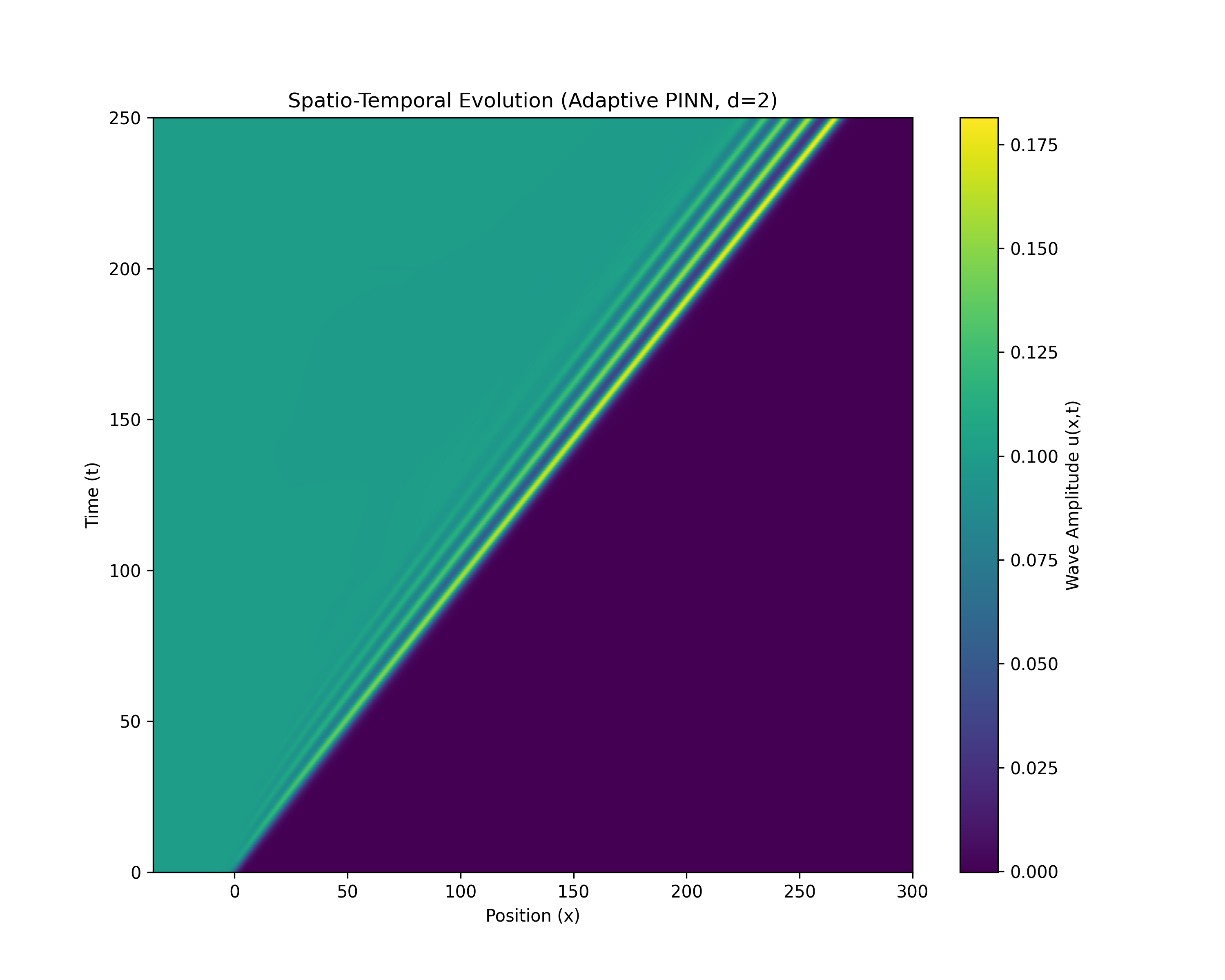}
    \caption{Spatio-temporal heatmap of the undular bore evolution predicted by the Adaptive PINN for $d=2$.}
    \label{fig:adaptive_pinn_steep}
\end{figure}

\begin{figure}[htbp]
\centering
    \includegraphics[width=0.8\textwidth]{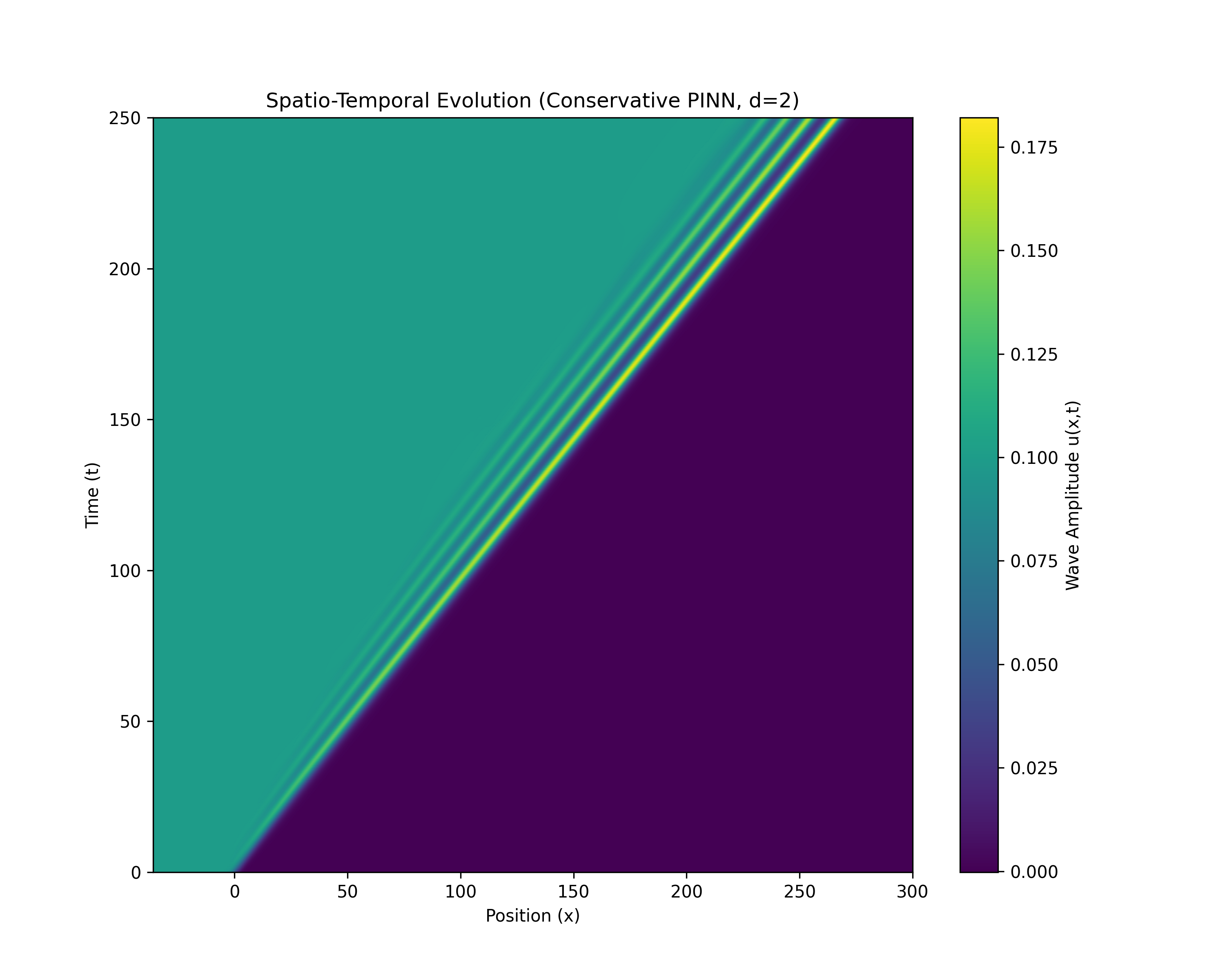}
    \caption{Spatio-temporal heatmap of the undular bore evolution predicted by the Conservative PINN for $d=2$.}
    \label{fig:conservative_pinn_steep}
\end{figure}

\begin{figure}[htbp]
\centering
    \includegraphics[width=0.8\textwidth]{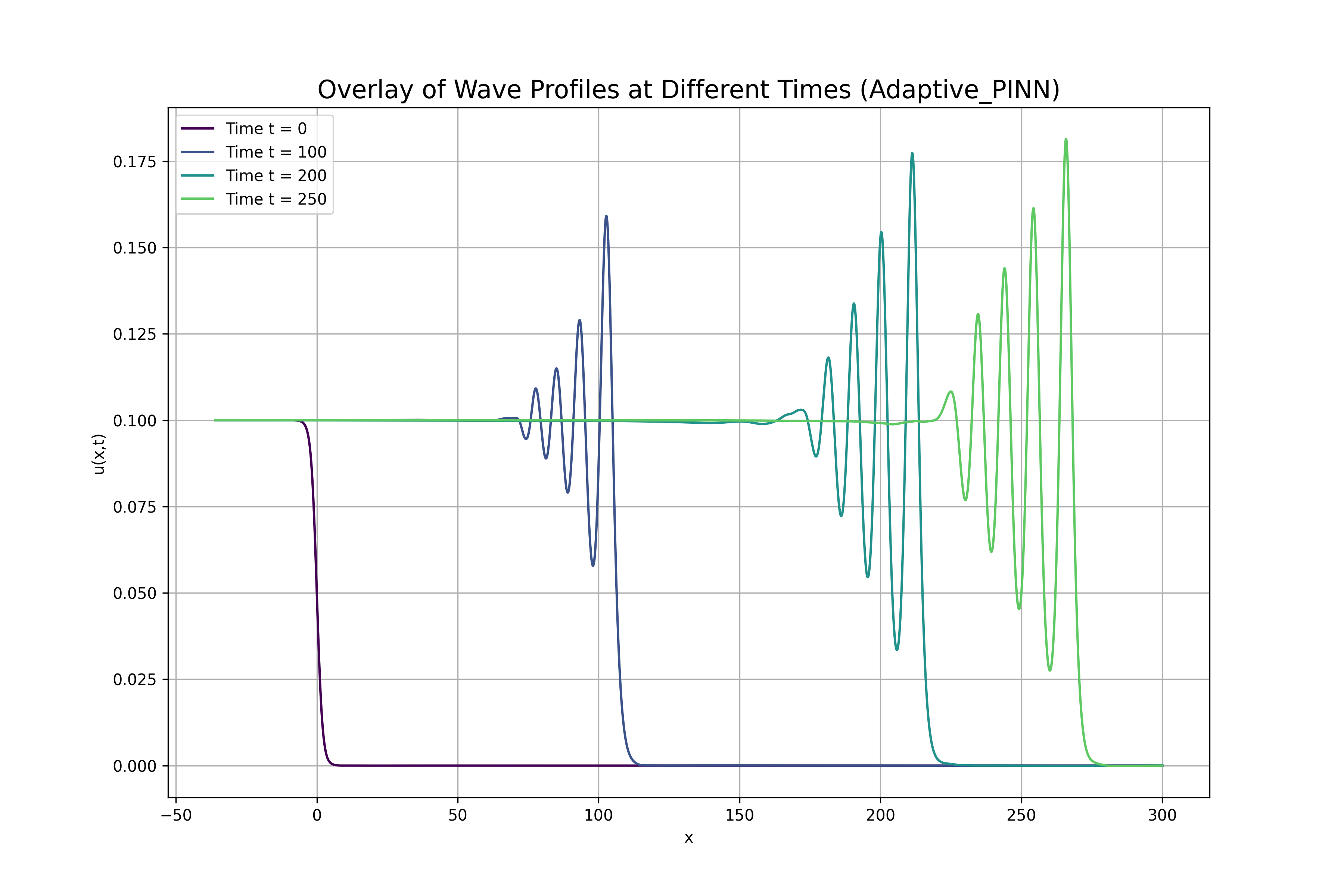}
    \caption{Overlay of wave profiles at key times ($t=0, 100, 200, 250$) for the Adaptive PINN solution.}
    \label{fig:adaptive_snapshots}
\end{figure}

\begin{figure}[htbp]
\centering
    \includegraphics[width=0.8\textwidth]{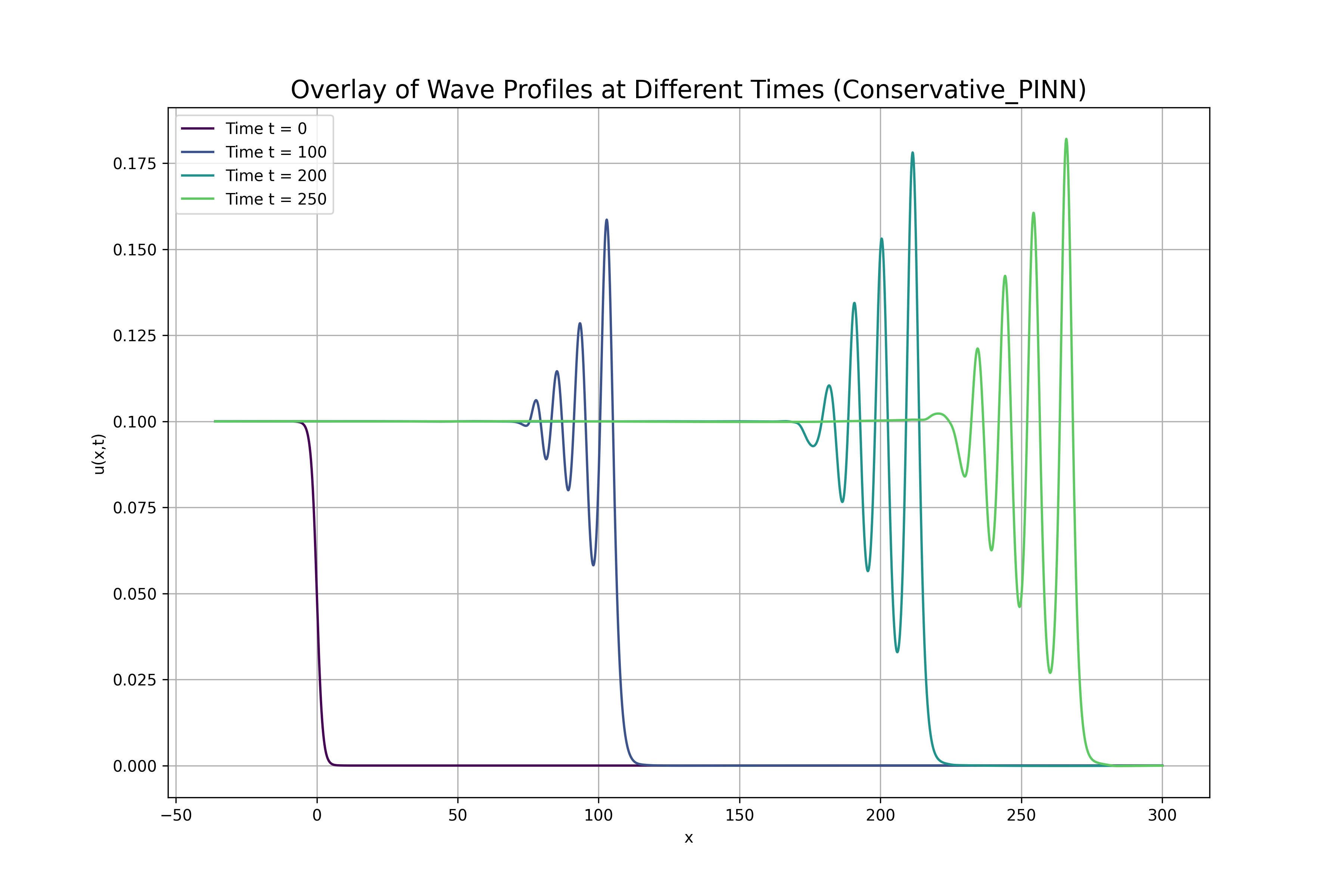}
    \caption{Overlay of wave profiles at key times for the Conservative PINN solution.}
    \label{fig:conservative_snapshots}
\end{figure}

\begin{figure}[htbp]
\centering
    \includegraphics[width=0.8\textwidth]{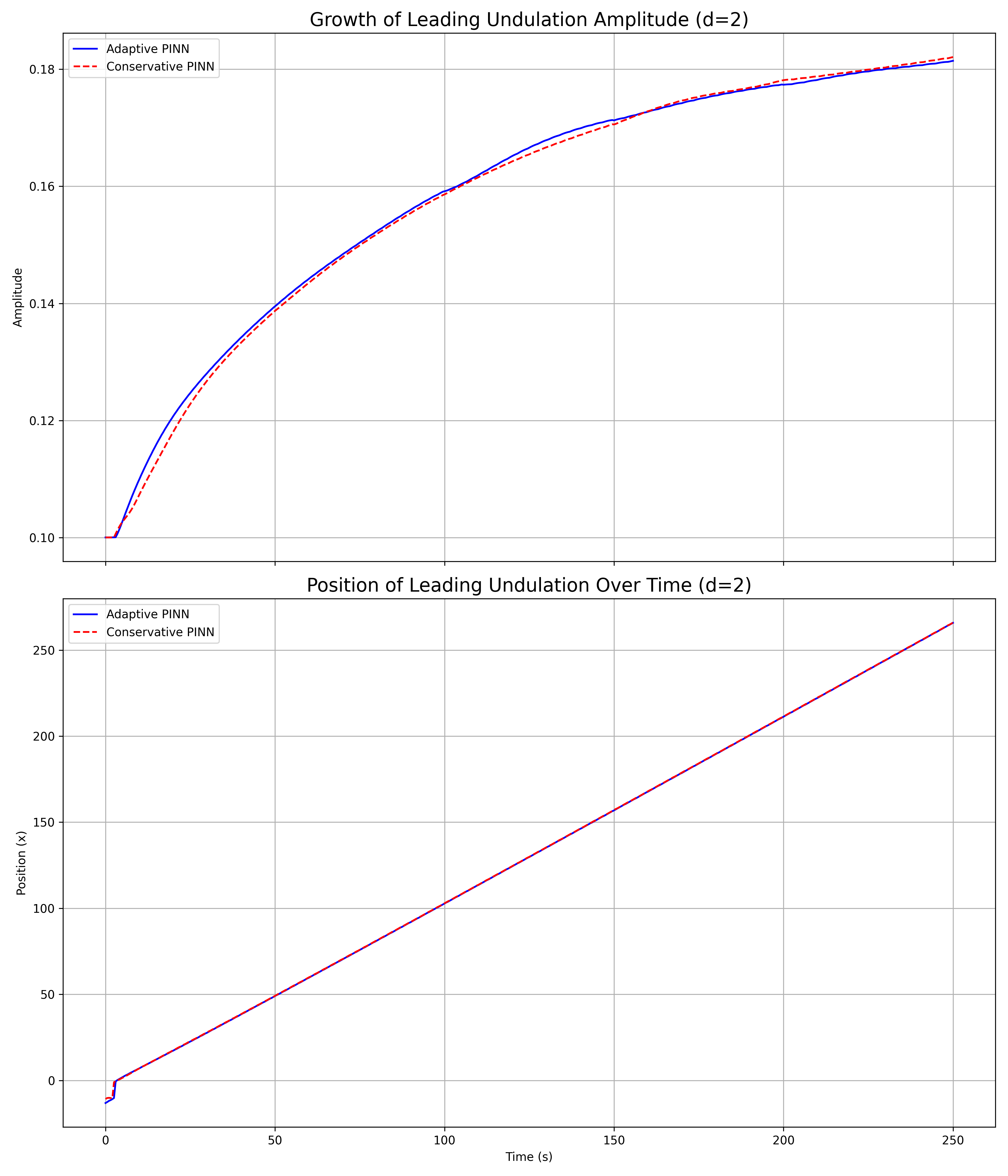}
    \caption{Growth of the leading undulation amplitude (top) and its position over time (bottom) for both methods, compared to the expected trajectory.}
    \label{fig:leading_wave_steep}
\end{figure}

The steep slope case ($d = 2$) presents a more challenging test. Results are quantified in Tables~\ref{tab:steep_slope_results}--\ref{tab:steep_slope_invariants}, with spatio-temporal evolution shown in Figures~\ref{fig:adaptive_pinn_steep}--\ref{fig:conservative_pinn_steep}, wave profiles in Figures~\ref{fig:adaptive_snapshots}--\ref{fig:conservative_snapshots}, and leading wave tracking in Figure~\ref{fig:leading_wave_steep}.

\begin{table}[htbp]
\centering
\caption{Final undulation properties at $t=250$ for the steep slope ($d=2$) case.}
\label{tab:steep_slope_results}
\begin{tabular}{lccc}
\toprule
\textbf{Method} & \textbf{Undulation} & \textbf{Position} & \textbf{Amplitude} \\
\midrule
Haq \& Ali \cite{haq2009} & Leading & 265.9 & 0.1820 \\
Adaptive PINN & Wave 1 & 265.8 & 0.1814 \\
Conservative PINN & Wave 1 & 266.0 & 0.1821 \\
Haq \& Ali \cite{haq2009} & Second & 254.2 & 0.1621 \\
Adaptive PINN & Wave 2 & 254.3 & 0.1614 \\
Conservative PINN & Wave 2 & 254.3 & 0.1606 \\
Haq \& Ali \cite{haq2009} & Third & 244.1 & 0.1446 \\
Adaptive PINN & Wave 3 & 244.0 & 0.1440 \\
Conservative PINN & Wave 3 & 244.2 & 0.1423 \\
\bottomrule
\end{tabular}
\end{table}

\begin{table}[htbp]
\centering
\caption{Leading wave amplitude at key times for the $d=2$ case.}
\label{tab:steep_slope_invariants}
\begin{tabular}{ccc}
\toprule
\textbf{Time} & \textbf{Adaptive PINN} & \textbf{Conservative PINN} \\
\midrule
0 & 0.1000 & 0.1000 \\
50 & 0.1395 & 0.1387 \\
100 & 0.1592 & 0.1586 \\
150 & 0.1713 & 0.1706 \\
200 & 0.1774 & 0.1781 \\
250 & 0.1814 & 0.1821 \\
\bottomrule
\end{tabular}
\end{table}

Causal training was essential for stable solutions to $t = 250$, successfully mitigating error accumulation in long-time integration.

Both approaches demonstrate remarkable capability for this complex, long-time evolution, but the Conservative PINN achieves a slight advantage in the most critical feature: the leading undulation amplitude. Tables~\ref{tab:steep_slope_invariants} and Figure~\ref{fig:leading_wave_steep} show that at $t = 250$, the Conservative prediction (0.1821) nearly matches the reference value (0.1820), while the Adaptive PINN slightly under-predicts it (0.1814).

This superior performance is also evident in the wave profiles (Figures~\ref{fig:adaptive_snapshots} and \ref{fig:conservative_snapshots}) and the properties of the leading waves in Table~\ref{tab:steep_slope_results}. The Conservative PINN more accurately captures the amplitude of the first wave and maintains a phase (position) error of less than 0.05\% for the leading wave. While both methods show some small deviations for subsequent waves, the Conservative approach consistently produces a wave train that is visually and quantitatively closer to the reference solution.

The conservation law enforcement can provide a great deal of benefit in terms of numerical stability when it comes to avoiding accumulation of errors during prolonged simulations. The hard constraints of conservation laws on mass, momentum and energy are extremely effective at enforcing regularization and maintaining the physics of the problem, resulting in a stable solution for the full duration of the 250 second simulation. Although the Adaptive PINN has performed well, the conservative PINN has provided the advantage of being physically grounded in a way that is best suited for extreme time scales.

\subsection{Overall Discussion and Comparative Analysis}\label{subsec:discussion}
We have demonstrated strong problem-dependent performance differences among the enhanced PINN approaches:

\textbf{Conservative PINN advantages.} Conservative PINN models demonstrate superior long-term stability and structure preservation (single solitary wave; undular bore) compared to their adaptive counterparts due to the inductive bias provided by hard constraint on conservation.

\textbf{Adaptive PINN advantages.} Adaptive PINN models are best suited to study transient, nonlinear interactions that occur between multiple solitary waves as seen in the two-soliton collision scenario. The adaptive nature of the model is advantageous because it provides flexibility through its self-adjusting loss weights to traverse a more complex optimization landscape of highly nonlinear dynamical systems than the constrained Conservative model can.

\textbf{Key finding: When conservation hurts optimization.} An important outcome of this research is the observation that the explicit inclusion of conservation principles (the basis of physics-informed learning) may hinder optimization. As shown with the two-soliton interaction example, the additional constraints placed upon the neural network will limit its ability to freely search for the optimal solution to the nonlinear dynamics involved in the interaction process. In fact, the rigid enforcement of conservation during the complex soliton interaction phase hinders the networks ability to learn the detailed dynamics of the soliton interaction by restricting its search for the optimal solution to poor local optima. The self-balancing mechanism of the Adaptive PINN allows it to temporarily reduce the importance of conservation during difficult phases of the interaction process so that it can explore the solution space to learn the detailed dynamics of the soliton interaction process and re-enforce conservation once the core dynamics are understood. Therefore, this finding suggests that the design of a PINN should be tailored to the specific physical phenomenon being studied rather than uniformly applied.

\textbf{Causal training efficacy.} Causal training was found to be crucial for long time integration. However, depending on whether the PINN model is designed to study the dynamics or stability of a physical phenomenon, the use of an appropriate PINN configuration (Adaptive for studying the dynamics and Conservative for studying the stability) results in the accuracy of the enhanced PINNs to be equivalent to those obtained using traditional high fidelity methods while providing all the mesh free benefits associated with PINNs and providing a unified method for both forward and inverse problems.

\textbf{Computational considerations.} Training time for our enhanced PINNs varies greatly based on the number of solitons involved in the simulation. For single soliton cases, training times ranged from 2-4 hours, for two-soliton cases, training times were between 8-12 hours, and for fully simulated undular bores, training times were typically between 15-20 hours (using an NVIDIA P100 GPU). Although this is longer than typical computational times associated with numerical methods used to solve similar one-dimensional problems (minutes to hours), PINNs are beneficial for solving inverse problems and complex geometrical problems where traditional numerical methods fail.

\section{Limitations and Future Work}\label{sec:limitations}

While our enhanced PINN approaches show significant promise, several limitations remain:

\textbf{Performance is problem-dependent,} requiring \emph{a priori} architecture choice (Adaptive vs. Conservative) based on expected physics, which may not always be obvious. Practitioners must understand their problem characteristics to select the appropriate approach.

\textbf{Computational cost} of training these models, especially with causal domain decomposition, exceeds traditional numerical methods for these 1D problems. For single PDE solutions, traditional methods remain more efficient.

\textbf{Hyperparameters,} particularly the conservation loss weight $\lambda_{\text{cons}}$, require manual tuning for each problem. We found that $\lambda_{\text{cons}}$ should be reduced for more nonlinear problems (from $10^{-4}$ for single soliton to $10^{-5}$ for two-soliton interaction).

\textbf{Limited scope:} This study focuses on 1D scenarios with available validation data. Performance on higher-dimensional problems or substantially different PDE types remains unexplored.

\textbf{Statistical analysis:} We report results from single training runs. Multiple runs with different initializations would provide error bars and confidence intervals.

These limitations motivate several research directions:

\textbf{Automated method selection frameworks} that choose optimal PINN configurations based on problem characteristics would significantly enhance applicability. Machine learning meta-models could predict which approach (Adaptive vs. Conservative) will perform best for a given problem.

\textbf{Advanced optimization algorithms} specifically developed for use in physics informed learning will reduce computation time and may significantly improve convergence; especially if they can be used to effectively enforce conservation constraints in a way that does not destabilize the optimization process. Optimization strategies such as adaptive learning rate scheduling, clipping of gradients or use of specially optimized algorithms for this type of problem could assist in reducing computation time and improving convergence.

\textbf{Adaptive collocation point sampling} could also improve the efficiency of the computations by concentrating the computational resources where the largest changes occur in the solution gradients.

Critical next steps include developing theoretical frameworks to predict when enforcement of conservation laws during optimization improves the optimization process and when it hinders it, extension of this work into high dimensional problems (inverse problems for parameter identification, coupled nonlinear systems and those involving sparse or noisy data), and utilization of the mesh-free nature of PINNs to their advantage in high dimensional problems.

\section{Conclusion}\label{sec:conclusion}

Our study presents a significant, yet underappreciated finding: Adding conservation laws explicitly to PINNs may compromise optimization for highly non-linear transient processes. The two-soliton interaction shows the added physical constraints hinder the neural network's adaptability and make the loss surface significantly harder to optimize for the collisions between two complex solitons, resulting in physically incorrect solutions. This calls into question one of the most basic assumptions in physics-informed learning, and thus requires the development of sophisticated optimization techniques that effectively embed physical constraints in a way that will not disrupt training stability.

Physics-Informed Neural Networks (PINNs) have very large error rates when solving the Regularized Long Wave (RLW) equation. Therefore we developed and evaluated two new enhanced PINN architectures: Adaptive PINN using self-balancing loss weights and Conservative PINN using explicit conservation law enforcement. Also, we employed state-of-the-art training techniques --- two-stage curriculum and causal time-marching --- to solve accurately the three extremely difficult phenomena of single soliton propagation, two-soliton interaction, and long time undular bore development.

Our experiments show that PINNs are problem dependent in terms of their success. The Adaptive PINN was the most successful architecture for all three types of problems, and uniquely succeeded in achieving superior accuracy and physical correctness when solving the complex problem of two-soliton interactions. The Conservative PINN had mixed results: It increased stability for single solitons and improved long-time undular bore predictions, but its performance decreased for highly non-linear two-soliton interactions, and it produced physically inconsistent solutions.

The Adaptive PINN has demonstrated itself to be the most general-purpose architecture for the solution of various RLW phenomenon. On the other hand, the Conservative PINN has shown itself to be the best choice for problems requiring long term stability. Our work also shows that if PINNs are properly configured they can produce solutions that are at least as accurate as the corresponding solutions obtained using traditional high-fidelity numerical methods, with errors of $O(10^{-5})$ being comparable to those achieved by established RBF-based meshfree methods \cite{haq2009} and finite-difference methods, while retaining the advantages of meshfree methods for solving problems with complex geometries and inverse problems.

We believe our problem-dependent efficacies offer practical guidance: adaptive approaches for problems with complex transient behavior, and conservative approaches for problems with long-term stability. Future research directions include: (1) Developing automated method selection frameworks that predict the optimal configuration of a PINN based on the characteristics of a given problem; (2) Developing better optimization algorithms that robustly handle conservation constraints without destabilizing the training process; (3) Extending these studies to higher-dimensional problems where PINNs' advantages over traditional methods are most pronounced; and (4) Developing theoretical frameworks to predict whether enforcing conservation laws will help or hinder optimization, moving beyond purely empirical testing to develop principles for designing PINNs.


\begin{thebibliography}{99}
\bibitem{wazwaz2010} Wazwaz, A. M. (2010). Nonlinear partial differential equations. In \textit{Partial differential equations and solitary waves theory} (pp. 285-351). Berlin, Heidelberg: Springer Berlin Heidelberg.
\bibitem{debnath2005} Debnath, L. (2005). \textit{Nonlinear partial differential equations for scientists and engineers}. Boston, MA: Birkh\"{a}user Boston.
\bibitem{evans2022} Evans, L. C. (2022). \textit{Partial differential equations} (Vol. 19). American Mathematical Society.
\bibitem{leveque2007} LeVeque, R. J. (2007). \textit{Finite difference methods for ordinary and partial differential equations: steady-state and time-dependent problems}. Society for Industrial and Applied Mathematics.
\bibitem{anderson2002} Anderson, J. D. (2002). \textit{Computational fluid dynamics: the basics with applications} (pp. 86-90). New York: McGraw-Hill.
\bibitem{peregrine1966} Peregrine, D. H. (1966). Calculations of the development of an undular bore. \textit{Journal of Fluid Mechanics}, 25(2), 321-330.
\bibitem{benjamin1972} Benjamin, T. B., Bona, J. L., \& Mahony, J. J. (1972). Model equations for long waves in nonlinear dispersive systems. \textit{Philosophical Transactions of the Royal Society of London. Series A, Mathematical and Physical Sciences}, 272(1220), 47-78.
\bibitem{baskan2023} Baskan, A. (2023). A novel outlook to the an alternative equation for modelling shallow water wave: Regularised Long Wave (RLW) equation. \textit{Indian Journal of Pure and Applied Mathematics}, 54(1), 133-145.
\bibitem{shivamoggi1986} Shivamoggi, B. K. (1986). A symmetrical regularized long-wave equation for shallow-water waves. \textit{Physics of Fluids}, 29(3), 890-891.
\bibitem{seyler1984} Seyler, C. E., \& Fenstermacher, D. L. (1984). A symmetric regularized-long-wave equation. \textit{Physics of Fluids}, 27(1), 4-7.
\bibitem{bhowmik2019} Bhowmik, S. K., \& Karakoc, S. B. (2019). Numerical approximation of the generalized regularized long wave equation using Petrov--Galerkin finite element method. \textit{Numerical Methods for Partial Differential Equations}, 35(6), 2236-2257.
\bibitem{khan2025} Khan, H., Tamsir, M., Singh, M., Msmali, A. H., \& Meetei, M. Z. (2025). Numerical approximation of the time-fractional regularized long-wave equation emerging in ion acoustic waves in plasma. \textit{AIMS Mathematics}, 10(3), 5651-5670.
\bibitem{dehghan2011} Dehghan, M., \& Salehi, R. (2011). The solitary wave solution of the two-dimensional regularized long-wave equation in fluids and plasmas. \textit{Computer Physics Communications}, 182(12), 2540-2549.
\bibitem{kaya2003} Kaya, D., \& El-Sayed, S. M. (2003). An application of the decomposition method for the generalized KdV and RLW equations. \textit{Chaos, Solitons \& Fractals}, 17(5), 869-877.
\bibitem{guo2019} Guo, C., Li, F., Zhang, W., \& Luo, Y. (2019). A conservative numerical scheme for Rosenau-RLW equation based on multiple integral finite volume method. \textit{Boundary Value Problems}, 2019(1), 168.
\bibitem{alzaid2013} Al-Zaid, N. A., Bakodah, H. O., \& Hendi, F. A. (2013). Numerical solutions of the regularized long-wave (RLW) equation using new modification of Laplace-decomposition method. \textit{Advances in Pure Mathematics}, 3(1), 159-163.
\bibitem{redouane2023} Redouane, K. L., Arar, N., Ben Makhlouf, A., \& Alhashash, A. (2023). A higher-order improved Runge--Kutta method and cubic B-spline approximation for the one-dimensional nonlinear RLW equation. \textit{Mathematical Problems in Engineering}, 2023(1), 4753873.
\bibitem{bota2014} Bota, C., \& Caruntu, B. (2014). Approximate analytical solutions of the regularized long wave equation using the optimal homotopy perturbation method. \textit{The Scientific World Journal}, 2014(1), 721865.
\bibitem{kutluay2006} Kutluay, S., \& Esen, A. (2006). A finite difference solution of the regularized long-wave equation. \textit{Mathematical Problems in Engineering}, 2006(1), 085743.
\bibitem{pan2015} Pan, X., Che, H., \& Wang, Y. (2015). A high-accuracy compact conservative scheme for generalized regularized long-wave equation. \textit{Boundary Value Problems}, 2015(1), 141.
\bibitem{ghiloufi2020} Ghiloufi, A., Rahmeni, M., \& Omrani, K. (2020). Convergence of two conservative high-order accurate difference schemes for the generalized Rosenau--Kawahara-RLW equation. \textit{Engineering with Computers}, 36(2), 617-632.
\bibitem{alotaibi2023} Alotaibi, N., \& Alzubaidi, H. (2023). Solitary wave solutions of the MRLW equation using a spatial five-point stencil of finite difference approximation. \textit{Journal of Umm Al-Qura University for Applied Sciences}, 9(3), 221-229.
\bibitem{mei2014} Mei, L., Gao, Y., \& Chen, Z. (2014). A Galerkin finite element method for numerical solutions of the modified regularized long wave equation. \textit{Abstract and Applied Analysis}, 2014(1), 438289.
\bibitem{liu2013} Liu, Y., Li, H., Du, Y., \& Wang, J. (2013). Explicit multistep mixed finite element method for RLW equation. \textit{Abstract and Applied Analysis}, 2013(1), 768976.
\bibitem{guo1988} Guo, B. Y., \& Cao, W. M. (1988). The Fourier pseudospectral method with a restrain operator for the RLW equation. \textit{Journal of Computational Physics}, 74(1), 110-126.
\bibitem{kang2015} Kang, X., Cheng, K., \& Guo, C. (2015). A second-order Fourier pseudospectral method for the generalized regularized long wave equation. \textit{Advances in Difference Equations}, 2015(1), 339.
\bibitem{hong2020} Hong, Q., Wang, Y., \& Gong, Y. (2020). Optimal error estimate of two linear and momentum-preserving Fourier pseudo-spectral schemes for the RLW equation. \textit{Numerical Methods for Partial Differential Equations}, 36(2), 394-417.
\bibitem{haq2009} Haq, S., \& Ali, A. (2009). A meshfree method for the numerical solution of the RLW equation. \textit{Journal of Computational and Applied Mathematics}, 223(2), 997-1012.
\bibitem{dag2010} Dag, I., \& Dereli, Y. (2010). Numerical solution of RLW equation using radial basis functions. \textit{International Journal of Computer Mathematics}, 87(1), 63-76.
\bibitem{dehghan2021} Dehghan, M., \& Shafieeabyaneh, N. (2021). Local radial basis function--finite-difference method to simulate some models in the nonlinear wave phenomena: regularized long-wave and extended Fisher--Kolmogorov equations. \textit{Engineering with Computers}, 37(2), 1159-1179.
\bibitem{oruc2025} Oruc, O. (2025). An extrapolated second-order BDF combined with local meshfree radial point interpolation method for numerical simulation of multi-dimensional regularized long wave equation emerging in fluids. \textit{Journal of Scientific Computing}, 104(2), 44.
\bibitem{raissi2019} Raissi, M., Perdikaris, P., \& Karniadakis, G. E. (2019). Physics-informed neural networks: A deep learning framework for solving forward and inverse problems involving nonlinear partial differential equations. \textit{Journal of Computational Physics}, 378, 686-707.
\bibitem{karniadakis2021} Karniadakis, G. E., Kevrekidis, I. G., Lu, L., Perdikaris, P., Wang, S., \& Yang, L. (2021). Physics-informed machine learning. \textit{Nature Reviews Physics}, 3(6), 422-440.
\bibitem{lu2021} Lu, L., Meng, X., Mao, Z., \& Karniadakis, G. E. (2021). DeepXDE: A deep learning library for solving differential equations. \textit{SIAM Review}, 63(1), 208-228.
\bibitem{lu2021a} Lu, L., Jin, P., Pang, G., Zhang, Z., \& Karniadakis, G. E. (2021). Learning nonlinear operators via DeepONet based on the universal approximation theorem of operators. \textit{Nature Machine Intelligence}, 3(3), 218-229.
\bibitem{nguyen2023} Nguyen, V. G., Jung, S., An, H., \& Lee, G. (2023). Exploring the power of physics-informed neural networks for accurate and efficient solutions to 1D shallow water equations. \textit{Journal of Korea Water Resources Association}, 56(12), 939-953.
\bibitem{li2024} Li, Y., Sun, Q., Wei, J., \& Huang, C. (2024). An improved PINN algorithm for shallow water equations driven by deep learning. \textit{Symmetry}, 16(10), 1376.
\bibitem{qi2024} Qi, X., de Almeida, G. A., \& Maldonado, S. (2024). Physics-informed neural networks for solving flow problems modeled by the 2D shallow water equations without labeled data. \textit{Journal of Hydrology}, 636, 131263.
\bibitem{mcclenny2023} McClenny, L. D., \& Braga-Neto, U. M. (2023). Self-adaptive physics-informed neural networks. \textit{Journal of Computational Physics}, 474, 111722.
\bibitem{zhu2025} Zhu, P., Liu, Z., Xu, Z., \& Lv, J. (2025). An adaptive weight physics-informed neural network for vortex-induced vibration problems. \textit{Buildings}, 15(9), 1533.
\bibitem{hu2023} Hu, Z., Jagtap, A. D., Karniadakis, G. E., \& Kawaguchi, K. (2023). Augmented physics-informed neural networks (APINNs): A gating network-based soft domain decomposition methodology. \textit{Engineering Applications of Artificial Intelligence}, 126, 107183.
\bibitem{luo2025} Luo, D., Jo, S. H., \& Kim, T. (2025). Progressive domain decomposition for efficient training of physics-informed neural network. \textit{Mathematics}, 13(9), 1515.
\bibitem{munzer2022} Munzer, M., \& Bard, C. (2022). A curriculum-training-based strategy for distributing collocation points during physics-informed neural network training. \textit{arXiv preprint arXiv:2211.11396}.
\bibitem{guo2025} Guo, Y., Fu, Z., Min, J., Lin, S., Liu, X., Rashed, Y. F., \& Zhuang, X. (2025). Long-term simulation of physical and mechanical behaviors using curriculum-transfer-learning based physics-informed neural networks. \textit{arXiv preprint arXiv:2502.07325}.
\bibitem{mohammadi2025} Mohammadi, N., Abbaszadeh, M., Dehghan, M., \& Heitzinger, C. (2025). Parameter identification of shallow water waves using the generalized equal width equation and physics-informed neural networks: a conservative approximation scheme. \textit{Nonlinear Dynamics}, 113(7), 6491-6516.
\bibitem{nakamula2025} Nakamula, A., Obuse, K., Sawado, N., Shimasaki, K., Shimazaki, Y., Suzuki, Y., \& Toda, K. (2025). Discovery of quasi-integrable equations from traveling-wave data using the physics-informed neural networks. \textit{Physica Scripta}, 100(5), 056012.
\bibitem{moseley2020} Moseley, B., Markham, A., \& Nissen-Meyer, T. (2020). Solving the wave equation with physics-informed deep learning. \textit{arXiv preprint arXiv:2006.11894}.
\bibitem{ortiz2025} Ortiz Ortiz, R. D., Marin Ramirez, A. M., \& Ortiz Marin, M. A. (2025). Physics-informed neural networks and Fourier methods for the generalized Korteweg--de Vries equation. \textit{Mathematics}, 13(9), 1521.
\bibitem{finch2025} Finch, L., Dai, W., \& Bora, A. (2025). An artificial neural network method for simulating soliton propagation based on the Rosenau-KdV-RLW equation on unbounded domains. \textit{Mathematics}, 13(7), 1139.
\bibitem{demir2024} Demir, K. T., Logemann, K., \& Greenberg, D. S. (2024). Closed-boundary reflections of shallow water waves as an open challenge for physics-informed neural networks. \textit{Mathematics}, 12(21), 3315.
\bibitem{delamata2023} de la Mata, F. F., Gijon, A., Molina-Solana, M., \& Gomez-Romero, J. (2023). Physics-informed neural networks for data-driven simulation: Advantages, limitations, and opportunities. \textit{Physica A: Statistical Mechanics and its Applications}, 610, 128415.
\bibitem{kaplarevic2023} Kaplarevic-Malisic, A., Andrijevic, B., Bojovic, F., Nikolic, S., Krstic, L., Stojanovic, B., \& Ivanovic, M. (2023). Identifying optimal architectures of physics-informed neural networks by evolutionary strategy. \textit{Applied Soft Computing}, 146, 110646.
\bibitem{zeng2025} Zeng, W., Lu, R., \& Liu, T. (2025). CLINN: Conservation law informed neural network for approximating discontinuous solutions. \textit{arXiv preprint arXiv:2509.02091}.
\bibitem{baez2024} Baez, A., Zhang, W., Ma, Z., Das, S., Nguyen, L. M., \& Daniel, L. (2024). Guaranteeing conservation laws with projection in physics-informed neural networks. \textit{arXiv preprint arXiv:2410.17445}.
\bibitem{zhou2025} Zhou, C., Chen, J., Yang, Z., \& Png, C. E. (2025). Dual-balancing for physics-informed neural networks. \textit{arXiv preprint arXiv:2505.11117}.
\bibitem{wang2024} Wang, Y., \& Yang, S. (2024). Coupled integral PINN for conservation law. \textit{arXiv preprint arXiv:2411.11276}.
\bibitem{sundar2025} Sundar, R., Lucor, D., \& Sarkar, S. (2025). Sequential learning based PINNs to overcome temporal domain complexities in unsteady flow past flapping wings. \textit{arXiv preprint arXiv:2503.15679}.
\bibitem{roy2024} Roy, P., \& Castonguay, S. T. (2024). Exact enforcement of temporal continuity in sequential physics-informed neural networks. \textit{Computer Methods in Applied Mechanics and Engineering}, 430, 117197.
\bibitem{meng2020} Meng, X., Li, Z., Zhang, D., \& Karniadakis, G. E. (2020). PPINN: Parareal physics-informed neural network for time-dependent PDEs. \textit{Computer Methods in Applied Mechanics and Engineering}, 370, 113250.
\end{thebibliography}
\end{document}